\documentclass{article}
\PassOptionsToPackage{numbers, compress}{natbib}
\usepackage[utf8]{inputenc} 
\usepackage[T1]{fontenc}    
\usepackage{url}            
\usepackage{amsfonts}       
\usepackage{nicefrac}       
\usepackage{microtype}
\usepackage{graphicx}
\usepackage{booktabs} 
\usepackage{hyperref}
\usepackage{ifthen}
\usepackage{enumitem}

\usepackage{natbib}
\setcitestyle{authoryear,open={(},close={)}}

\usepackage[preprint]{neurips2021}

\usepackage[dvipsnames]{xcolor}
\usepackage{graphicx}
\usepackage{float}
\usepackage{caption} 
\usepackage{subcaption}
\usepackage{overpic}
\usepackage{mathtools}
\usepackage{tabularx}
\usepackage{amsmath,amssymb}
\usepackage{xspace}
\usepackage{adjustbox}
\usepackage{multirow}
\usepackage{wrapfig}
\usepackage{tikz}
\usepackage[nameinlink]{cleveref}
\usepackage[acronym]{glossaries}
\usepackage{placeins}
\usepackage{amssymb}
\usepackage{pifont}
\newcommand{\cmark}{\ding{51}}%
\newcommand{\xmark}{\ding{55}}%

\newtheorem{theorem}{Theorem}[section]

\usepackage{scalerel}
\usepackage{mleftright}
\usepackage{dsfont}
\usepackage{bm}
\newcommand\scalesym[2]{\hstretch{#1}{\vstretch{#1}{#2}}}

\glsdisablehyper{}
\newacronym{ERM}{ERM}{empirical risk minimisation}
\newacronym{MAML}{MAML}{model-agnostic meta-learning}
\newacronym{SGD}{SGD}{stochastic gradient descent}

\DeclareMathOperator{\E}{{}\mathbb{E}}

\DeclareMathOperator*{\argmin}{arg\,min} 

\renewcommand{\~}{\,\scalesym{0.9}{\sim}\,}          

\newcommand{\bg}{\bar{g}}
\newcommand{\bG}{\bar{G}}

\newcommand{\tg}{\widetilde{g}}
\newcommand{\g}{g}
\newcommand{\G}{{G}}
\newcommand{\pG}{\widehat{G}}
\newcommand{\pg}{\widehat{g}}

\renewcommand{\L}{\ensuremath{\mathcal{L}}}

\newcommand{\D}{\ensuremath{\mathcal{D}}}
\renewcommand{\d}{\ensuremath{d}}

\newcommand{\Da}{\ensuremath{{\color{c2}\D_1}}}
\newcommand{\Db}{\ensuremath{{\color{c3}\D_2}}}
\newcommand{\Dc}{\ensuremath{{\color{c1}\D_3}}}

\newcommand{\std}[1]{(\textit{\scriptsize$\pm$\scriptsize #1})}

\newcommand*\circled[1]{\tikz[baseline=(char.base)]{%
            \node[shape=circle,draw,inner sep=1pt] (char) {#1};}}

\makeatletter
\newcommand{\subalign}[1]{%
  \vcenter{%
    \Let@ \restore@math@cr \default@tag
    \baselineskip\fontdimen10 \scriptfont\tw@
    \advance\baselineskip\fontdimen12 \scriptfont\tw@
    \lineskip\thr@@\fontdimen8 \scriptfont\thr@@
    \lineskiplimit\lineskip
    \ialign{\hfil$\m@th\scriptstyle##$&$\m@th\scriptstyle{}##$\hfil\crcr
      #1\crcr
    }%
  }%
}
\makeatother

\newcommand{\norm}[1]{\left\lVert#1\right\rVert}

\definecolor{c1}{HTML}{086788}
\definecolor{c2}{HTML}{BA553C}
\definecolor{c3}{HTML}{E8A321}
\definecolor{c4}{HTML}{5D8270}
\definecolor{c5}{HTML}{89B68C}

\usepackage{algorithm}
\usepackage{algorithmic}
\renewcommand{\algorithmiccomment}[1]{{\tt /\!\!/#1}}

\begin{document}
\title{Gradient Matching for Domain Generalization}
\author{%
Yuge Shi\thanks{Work done during internship at Facebook AI Research.}\\
University of Oxford \\
\texttt{yshi@robots.ox.ac.uk}
\And
Jeffrey Seely\\
Facebook Reality Labs\\
\texttt{jseely@fb.com}
\And
Philip H.S. Torr\\
University of Oxford\\
\texttt{philip.torr@eng.ox.ac.uk}
\And
N. Siddharth\\
The University of Edinburgh \\
\& The Alan Turing Institute\\
\texttt{n.siddharth@ed.ac.uk}
\And
Awni Hannun\\
Facebook AI Research\\
\texttt{awni@fb.com}
\And
Nicolas Usunier\\
Facebook AI Research\\
\texttt{usunier@fb.com}
\And
Gabriel Synnaeve\\
Facebook AI Research\\
\texttt{gab@fb.com}
}
 \maketitle

\setlength{\tabcolsep}{4pt}
\vspace{-15pt}
\begin{abstract}
Machine learning systems typically assume that the distributions of training and test sets match closely. 
However, a critical requirement of such systems in the real world is their ability to generalize to unseen domains. 
Here, we propose an \emph{inter-domain gradient matching} objective that targets domain generalization by maximizing the inner product between gradients from different domains.
Since direct optimization of the gradient inner product can be computationally prohibitive --- it requires computation of second-order derivatives –-- we derive a simpler first-order algorithm named Fish that approximates its optimization. 
We perform experiments on the \textsc{Wilds} benchmark, which captures distribution shift in the real world, as well as the \textsc{DomainBed} benchmark that focuses more on synthetic-to-real transfer. 
Our method produces competitive results on both benchmarks, demonstrating its effectiveness across a wide range of domain generalization tasks.
Code is available at \href{https://github.com/YugeTen/fish}{\texttt{https://github.com/YugeTen/fish}}.
\end{abstract}

\section{Introduction}
\begin{wrapfigure}[12]{r}{0.45\textwidth}
    \vspace{-14pt}
    \includegraphics[width=1.0\linewidth, trim={0.28cm 0.15cm 0.18cm 0.14cm}, clip]{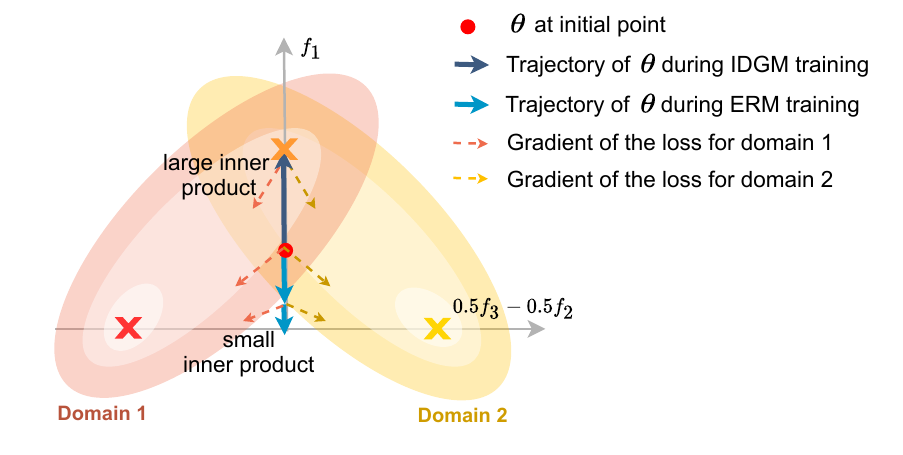}
    \caption{Isometric projection of training with ERM (blue) vs. our IDGM objective (dark blue), using data from \Cref{fig:toy}.}
    \label{fig:introfig}
\end{wrapfigure}

The goal of domain generalization is to train models that performs well on unseen, out-of-distribution data, which is crucial in practice for model deployment in the wild.
This seemingly difficult task is made possible by the presence of multiple distributions/domains at train time. 
As we have seen in past work \citep{irm, deepcoral, ganin2016domain}, a key aspect of domain generalization is to learn from features that remain \emph{invariant} across multiple domains, while ignoring those that are \emph{spuriously correlated} to label information (as defined in \citet{unbiasedlook, stock2017convnets}).
Consider, for example, a model that is built to distinguish between cows and camels using photos collected in nature under different climates.
Since CNNs are known to have a bias towards texture \citep{texture1, texture2}, if we simply try to minimize the average loss across different domains, the classifier is prone to spuriously correlate ``cow'' with grass and ``camels'' with desert, and predict the species using only the background.
Such a classifier can be rendered useless when the animals are placed indoors or in a zoo.
However, if the model could recognize that while the landscapes change with climate, the biological characteristics of the animals (e.g. humps, neck lengths) remain invariant and use those features to determine the species, we have a much better chance at generalizing to unseen domains.

Similar intuitions have already motivated several approaches that consider learning ``invariances'' accross domains as the main challenge of domain generalization.
%
%
Most of these work focuses on learning \emph{invariant features}, for instance domain adversarial neural networks \citep{ganin2016domain}, CORAL \citep{sun2016deep} and MMD for domain generalization \citep{li2018domain}.
Different from previous approaches, invariant risk minimization \citep{irm} proposes to learn intermediate features such that we have \emph{invariant predictor} (when optimal) across different domains.

In this paper, we propose an inter-domain gradient matching (IDGM) objective.
While our method also follows the invariance assumption, we are interested in learning a model with \emph{invariant gradient direction} for different domains.
Our IDGM objective augments the loss with an auxiliary term that maximizes the gradient inner product between domains, which encourages the alignment between the domain-specific gradients. By simultaneously minimizing the loss and matching the gradients, IDGM encourages the optimization paths to be the same for all domains, favouring invariant predictions. See \Cref{fig:introfig} for a visualization: given 2 domains, each containing one invariant feature (orange cross) and one spurious feature (yellow and red cross). While \gls{ERM} minimizes the average loss between these domains at the cost of learning spurious features only, IDGM maximizes the gradient inner product and is therefore able to focus on the invariant feature. Note that this plot is generated from an example, which we will describe in more details in \Cref{sec:ermbad}.

While the IDGM objective achieves the desirable learning dynamic in theory, naive optimization of the objective by gradient descent is computationally costly due to the second-order derivatives. Leveraging the theoretical analysis of Reptile, a meta-learning algorithm~\citep{reptile}, we propose to approximate the gradients of IDGM using a simple first-order algorithm, which we name Fish. Fish is simple to implement, computationally effective and as we show in our experiments, functionally similar to direct optimization of IDGM.

Our contribution is a simple but effective training algorithm for domain generalization, which exhibits state-of-the-art performance on 13 datasets from recent domain generalization benchmark \textsc{Wilds} \citep{wilds} and \textsc{DomainBed}. The strong performance of our method on a wide variety of datasets demonstrates that it is broadly applicable in different applications and subgenres of domain generalization problems. We also perform experiments to verify that our algorithm Fish does improve the normalized inter-domain gradient inner product, while this inner product decreases throughout training for \gls{ERM} baseline.

\section{Related Work}\label{sec:related}
\textbf{Domain Generalization} In domain generalization, the training data is sampled from one or many source domains, while the test data is sampled from a new target domain. In contrast to domain \emph{adaptation}, the learner does not have access to any data from the target domain (labeled or unlabeled) during train time \citep{quionero2009dataset}. 
In this paper we are interested in the scenario where multiple source domains are available, and the domain where the data comes from is known.
Further, \citet{wilds} defines the datasets where train and test has disjoint domains ``domain generalization'', and those where domains overlap between splits (but typically have different distributions) ``subpopulation shift''.
Following e.g., \citet{deepcoral}, in this paper we use domain generalization in a broader sense that encompasses the two categories.

We will now discuss the three main families of approaches to domain generalization: 
\begin{enumerate}[leftmargin=0.3cm]
    \item \textbf{Distributional Robustness (DRO)}: DRO approaches minimize the worst-case loss over a set of data distributions constructed from the training domains. \citet{rojas2015causal} proposed DRO to address \emph{covariate shift} \citep{gretton2008covariate,gretton2009covariate}, where $P(Y|X)$ remains constant across domains but $P(X)$ changes. Later work also studied \emph{subpopulation shift}, where the train and test distributions are mixtures of the same domains, but the mixture weights change between train and test \citep{hu2018does,sagawa2019distributionally};
    \item \textbf{Domain-invariant representation learning}: This family of approaches to domain generalization aims at learning high-level features that make domains statistically indistinguishable. Prediction is then based on these features only. The principle is motivated by a generalization error bound for unsupervised domain adaptation \citep{ben2010theory,ganin2016domain}, but the approach readily applies to domain generalization \citep{deepcoral,wilds}. Algorithms include penalising the domain-predictive power of the model \citep{ganin2016domain, wang2019learning, huang2020self}, matching mean and variance of feature distributions across domains \citep{sun2016deep}, learning useful representations by solving Jigsaw puzzles \citep{carlucci2019domain}  or using the maximum mean discrepancy \citep{gretton2006kernel} to match the feature distributions \citep{li2018domain}. 
    
    Similar to our approach, \citet{IGA} proposes IGA, which also adopts a gradient-alignment approach for domain generalization. The key difference between IGA and our IDGM objective is that IGA learns invariant features by minimizing the \emph{variance} of inter-domain gradients. Notably, IGA is completely identical to ERM when ERM is the optimal solution on every training domain, since the variances of the gradients will be zero. While they achieve the best performance on the training set, both IGA and ERM could completely fail when generalizing to unseen domains (see \Cref{sec:ermbad} for such an example). Our method, on the contrary, biases towards non-ERM solutions as long as the gradients are aligned, and is therefore able to avoid this issue.
    Additionally, in \citet{lopez-paz2017gradient} we also see the application of gradient-alignment, however in this case it is applied under the continual learning setting to determine whether a gradient update will increase the loss of the previous tasks.
    
    \item \textbf{Invariant Risk Minimization (IRM)}: IRM is proposed by \citet{irm}, which learns an intermediate representation such that the optimal classifiers (on top of this representation) of all domains are the same. The motivation is to exploit invariant causal effects between domains while reducing the effect of domain-specific spurious correlations. 
\end{enumerate}

Apart from these algorithms that are tailored for domain generalization, a well-studied baseline in this area is \gls{ERM}, which simply minimizes the average loss over training domains. Using vanilla \gls{ERM} is theoretically unfounded \citep{ermbad1, ermbad2, ermbad3} since ERM is guaranteed to work only when train and test distributions match.
Nonetheless, recent benchmarks suggest that ERM obtains strong performance in practice, in many case surpassing domain generalization algorithms \citep{deepcoral,wilds}. Our goal is to fill this gap, using an algorithm significantly simpler than previous approaches.

\paragraph{Connections to meta-learning}
There are close connections between meta-learning \citep{thrun1998learning} and (multi-source) domain adaptation. In fact, there are a few works in domain generalization that are inspired by the meta-learning principles, such as \citet{li2018learning, balaji2018metareg, li2019episodic, dou2019domain}.

Meta-learning aims at reducing the sample complexity of new, unseen tasks. A popular school of thinking in meta-learning is model agnostic meta-learning (MAML), first proposed in \citet{maml, andrychowicz2016learning}. The key idea is to backpropagate through gradient descent itself to learn representations that can be easily adapted to unseen tasks.
Our algorithmic solution is inspired by Reptile, a first-order approximation to MAML. However, our method has a fundamentally different goal, which is to exploit input-output correspondences that are invariant across domains. In contrast, meta-learning algorithms such as Reptile extract knowledge (e.g., input representations) that are useful to different tasks, but nothing has to be invariant across all tasks. 

While our motivation is the exploitation of invariant features, similarly to IRM, our inter-domain gradient matching principle is an alternative to specifying hard constraints on the desired invariants across tasks. The resulting algorithm is both simple and efficient as it is a combination of standard gradient computations and parameter updates.

\section{Methodology}
\label{sec:methodology}
\subsection{Goals}
Consider a training dataset $\D_{tr}$ consisting of $S$ domains $\D_{tr}=\{\D_1, \cdots,  \D_S\}$, where each domain $s$ is characterized by a dataset $\D_s:=\{(x^s_i,y^s_i)\}^{n_s}_{i=1}$ containing data drawn i.i.d. from some probability distribution.
Also consider a test dataset $\D_{te}$ consisting of $T$ domains $\D_{te}=\{\D_{S+1}, \cdots,  \D_{S+T}\}$, where $\D_{tr} \cap \D_{te} = \emptyset$.
The goal of domain generalization is to train a model with weights $\theta$ that generalizes well on the test dataset $\D_{te}$ such that:
\begin{align}
    \argmin_{\theta} \E_{\D \sim \D_{te}} \E_{(x,y) \sim \D}\left[ l((x,y); \theta) \right], \label{eq:obj}
\end{align}
where $l((x,y); \theta)$ is the loss of model $\theta$ evaluated on $(x,y)$.

%
%
A naive approach is to emply \gls{ERM}, which simply minimizes the average loss on $\D_{tr}$, ignoring the discrepancy between train and test domains:
\begin{align}
    \L_{\text{erm}}(\D_{tr};\theta) = \E_{\D\sim\D_{tr}} \E_{(x,y) \sim \D}\left[ l((x,y);\theta)\right]. \label{eq:erm}
\end{align}
The \gls{ERM} objective clearly does not exploit the input-output invariance across different domains in $\D_{tr}$ and could perform arbitrarily poorly on test data.
We demonstrate this with a simple linear example as described in the next section.

\begin{figure}[H]
    \centering
    \vspace{-4pt}
    \includegraphics[width=0.6\linewidth]{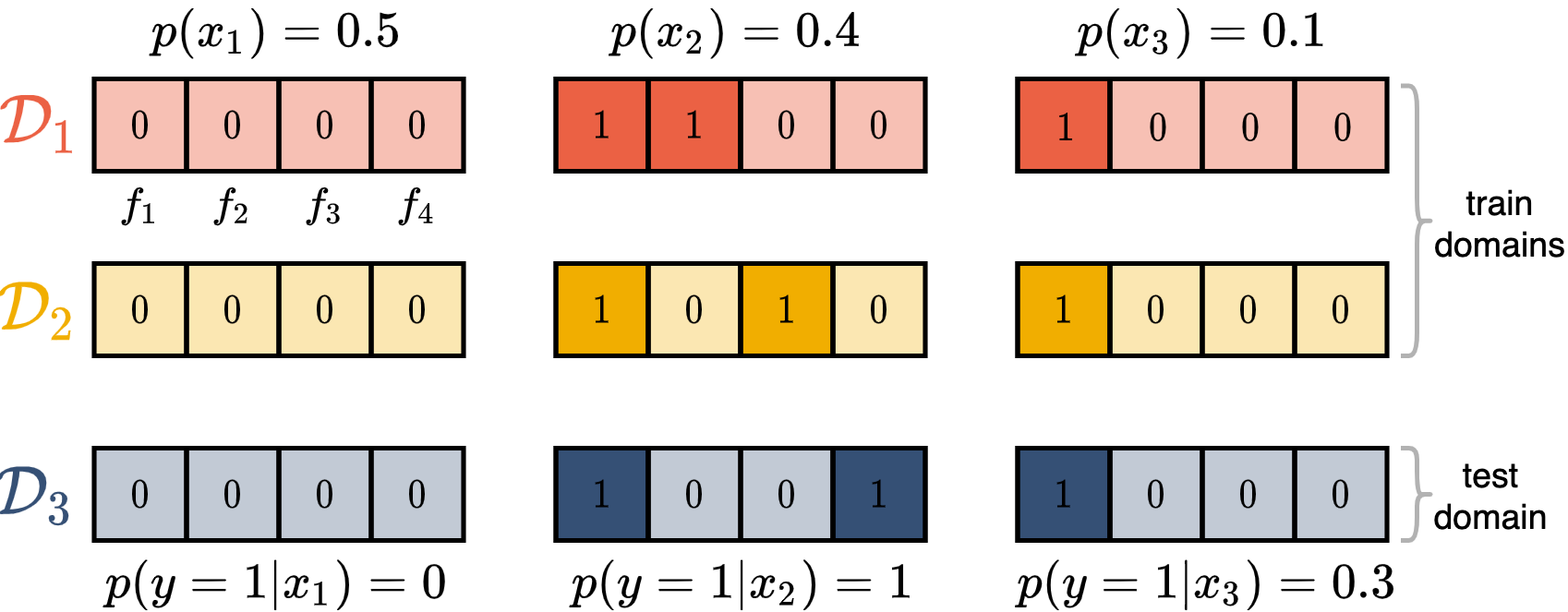}
    \caption{All domains contain 3 types of inputs $x_1, x_2$ and $x_3$, each depicted in one column.
    \textbf{\textit{1$^{st}$ col.}}: $x_1=[0,0,0,0]$, $y=0$, makes up for $50\%$ of each dataset;
    \textbf{\textit{2$^{nd}$ col.}}: $x_2$ changes for each domain, $y=1$ always. $40\%$ of each dataset;
    \textbf{\textit{3$^{rd}$ col.}}: $x_3=[1,0,0,0]$, $30\%$ of $y=1$ and $70\%$ of $y=0$. $10\%$ of each dataset.\label{fig:toy}}
    \vspace{-10pt}
\end{figure}

\subsection{The pitfall of ERM: a linear example}
\label{sec:ermbad}

Consider a binary classification setup where data~\((x, y) \in \mathbb{B}^4 \times \mathbb{B}\), and a data instance is denoted~\(x = [f_1, f_2, f_3, f_4], y\).
Training data spans two domains~$\{\Da, \Db\}$, and test data one~${\Dc}$.
The goal is to learn a linear model $Wx+b=y, W\in\mathbb{R}^4, b\in \mathbb{R}$ on the train data, such that the error on test data is minimized.
The setup and dataset of this example is illustrated in \Cref{fig:toy}.

As we can see in \Cref{fig:toy}, $f_1$ is the \emph{invariant feature} in this dataset, since the correlation between $f_1$ and $y$ is stable across different domains.
The relationships between $y$ and $f_2, f_3$ and $f_4$ changes for \Da, \Db, \Dc, making them the \emph{spurious features}.
Importantly, if we consider one domain only, the spurious feature is a more accurate indicator of the label than the invariant feature.
For instance, using $f_2$ to predict $y$ can give $97\%$ accuracy on \Da, while using $f_1$ only achieves $93\%$ accuracy.
However, predicting with $f_2$ on \Db~and \Dc~can at most reach $57\%$ accuracy, while $f_1$'s accuracy remains $93\%$ regardless of the domain.
\begin{table}[H]
\centering
\vspace{-4pt}
\caption{Performance comparison on the linear dataset.}
\scalebox{0.9}{
\begin{tabular}{ccccc}
    \toprule
    Method & train acc. & test acc. & $W$ & $b$ \\
    \midrule
    \gls{ERM} & $97\%$ & $57\%$ & $[2.8, 3.3, 3.3, 0.0]$ & $-2.7$ \\
    IDGM & $93\%$ & $93\%$ & $[0.4, 0.2, 0.2, 0.0]$ & $-0.4$ \\
    Fish & $93\%$ & $93\%$ & $[0.4, 0.2, 0.2, 0.0]$ & $-0.4$ \\
    \bottomrule
\end{tabular}\label{tab:toy_results}}
\end{table}

The performance of \gls{ERM} on this simple example is shown in \Cref{tab:toy_results} (first row).
From the trained parameters $W$ and $b$, we see that the model places most of its weights on spurious features $f_2$ and $f_3$.
While this achieves the highest train accuracy ($97\%$), the model cannot generalize to unseen domains and performs poorly on test accuracy ($57\%$).

\subsection{Inter-domain Gradient Matching (IDGM)} \label{sec:gradinnerprod}
%
To mitigate the problem with \gls{ERM}, we need an objective that learns from features that are invariant across domains.
Let us consider the case where the train dataset consists of $S=2$ domains $\D_{tr} = \{\D_1, \D_2\}$.
Given model $\theta$ and loss function $l$, the expected gradients for data in the two domains is expressed as
\begin{align}\label{eq:grads}
    \G_1 = \E_{\D_1}\frac{\partial l((x, y); \theta)}{\partial\theta}, \quad
    \G_2 = \E_{\D_2}\frac{\partial l((x, y); \theta)}{\partial\theta}.
\end{align}
The direction, and by extension, inner product of these gradients are of particular importance to our goal of learning invariant features. 
If $G_1$ and $G_2$ point in a similar direction, i.e. $\G_1 \cdot \G_2>0$, taking a gradient step along $\G_1$ or $\G_2$ improves the model's performance on both domains, indicating that the features learned by either gradient step are invariant across $\{\D_1, \D_2\}$.
This invariance cannot be guaranteed if $\G_1$ and $\G_2$ are pointing in opposite directions, i.e. $\G_1 \cdot \G_2 \leq 0$.

To exploit this observation, we propose to maximize the gradient inner product (GIP) to align the gradient direction across domains.
The intended effect is to find weights such that the input-output correspondence is as close as possible across domains.
We name our objective \emph{inter-domain gradient matching} (IDGM), and it is formed by subtracting the inner product of gradients between domains $\pG$ from the original \gls{ERM} objective.
For the general case where $S\geq 2$, we can write 
\begin{align}
\L_{\text{idgm}} &= \L_{\text{erm}}(\D_{tr}; \theta) -  \gamma \underbrace{\frac{2}{S(S-1)} \sum_{i,j\in S}^{i\neq j} \G_i\cdot \G_j}_{\text{GIP, denote as }\pG},\label{eq:obj_ip}
\end{align}
%
%
%
where $\gamma$ is the scaling term for $\pG$.
Note that GIP can be computed in linear time as $\pG = ||\sum_i G_i||^2 - \sum_i ||G_i||^2$ (ignoring the constant factor).
We can also compute the stochastic estimates of \Cref{eq:obj_ip} by replacing out the expectations over the entire dataset by minibatches.
%

We test this objective on our simple linear dataset, and report results in the second row of \Cref{tab:toy_results}.
Note that to avoid exploding gradient we use the normalized GIP during training.
The model has lower training accuracy compared to \gls{ERM} ($93\%$), however its accuracy remains the same on the test set, much higher than \gls{ERM}.
The trained weights $W$ reveal that the model assigns the largest weight to the invariant feature $f_1$, which is desirable.
The visualization in \Cref{fig:introfig} also confirms that by maximizing the gradient inner product, IDGM is able to focus on the feature that is common between domains, yielding better generalization performance than \gls{ERM}.
%

\subsection{Optimizing IDGM with Fish}
\label{sec:fish}
The proposed IDGM objective, although effective, requires computing the second-order derivative of the model's parameters due to the gradient inner product term, which can be computationally prohibitive.
To mitigate this, 
we propose a first-order algorithm named Fish\footnote{Following the convention of naming this style of algorithms after classes of vertebrates (animals with backbones).} that approximates the optimization of IDGM with inner-loop updates.
In \Cref{alg:fish} we present Fish. As a comparison, we also present direct optimization of IDGM using SGD in \Cref{alg:idgm}.
\begin{center}
\scalebox{0.9}{
\begin{minipage}[t]{0.53\linewidth}
\vspace{-30pt}
\begin{algorithm}[H]
    \begin{algorithmic}[1]
    \FOR{iterations = $1, 2, \cdots$}
        \STATE ${\widetilde{\theta}} \gets \theta$
        \FOR{$\D_i \in \texttt{permute}(\{\D_1, \D_2, \cdots, \D_S\})$}
            \STATE Sample batch  ${d}_i \sim \D_i$
            \STATE $\displaystyle \tg_i = \E_{\d_i}\left[\frac{\partial l((x,y);\widetilde{\theta})}{\partial{\widetilde{\theta}}}\right]$ \COMMENT{Grad wrt $\widetilde{\theta}$}
            \STATE Update ${\widetilde{\theta}} \gets \widetilde{\theta} - \alpha \tg_i$
        \ENDFOR
        \STATE $\vphantom{\bg =\displaystyle \frac{1}{S}\sum^S_{s=1} \g_s, \quad \pg = \underbrace{\frac{2}{S(S-1)}\sum_{i,j\in S}^{i\neq j} \g_i\cdot \g_j}}$
        \vspace{-11pt}
        \STATE Update $\theta \gets \theta + \displaystyle \epsilon(\widetilde{\theta}-\theta)$
    \ENDFOR
    \end{algorithmic}
\caption{Fish.}
\label{alg:fish}
\end{algorithm}
\end{minipage}\hspace{20pt}\begin{minipage}[t]{0.53\linewidth}
\vspace{-30pt}
\begin{algorithm}[H]
    \begin{algorithmic}[1]
    \FOR{iterations = $1, 2, \cdots$}
        \STATE ${\widetilde{\theta}} \gets \theta$
        \FOR{$\D_i \in \texttt{permute}(\{\D_1, \D_2, \cdots, \D_S\})$}
            \STATE Sample batch  ${d}_i \sim \D_i$
            \STATE $\displaystyle g_i = \E_{\d_i}\left[\frac{\partial l((x,y);\theta)}{\partial\theta}\right]$ \COMMENT{Grad wrt $\theta$}\vspace{3pt}
            \STATE
        \ENDFOR
        \vspace{-10pt}
        \STATE $\bg =\displaystyle \frac{1}{S}\sum^S_{s=1} \g_s, \quad \pg = \overbrace{\frac{2}{S(S-1)}\sum_{i,j\in S}^{i\neq j} \g_i\cdot \g_j}^{\text{GIP (batch)}}$
        \STATE Update $\theta \gets \theta - \epsilon\left(\bg-\gamma (\partial\pg/\partial \theta)\right)$
    \ENDFOR
    \end{algorithmic}
\caption{Direct optimization of IDGM.}
\label{alg:idgm}
\end{algorithm}
\end{minipage}}
\end{center}
%
Fish performs $S$ inner-loop (\textit{l3-l7}) update steps with learning rate $\alpha$ on a clone of the original model $\widetilde{\theta}$, and each update uses a minibatch $d_i$ from the domain selected in step $i$.
Subsequently, $\theta$ is updated by a weighted difference between the cloned model and the original model $\epsilon(\widetilde{\theta}-\theta)$.
%
%


To see why Fish is an approximation to directly optimizing IDGM, we can perform Taylor-series expansion on its update in \textit{l8}, \Cref{alg:fish}.
Doing so reveals two leading terms:
1) $\bg$: averaged gradients over inner-loop's minibatches (effectively the ERM gradient);
2) $\partial \pg/\partial \theta$: gradient of the minibatch version of GIP.
Observing \textit{l8} of \Cref{alg:idgm}, we see that $\bg$ and $\pg$ are actually the two gradient components used in direct optimization of IDGM.
Therefore, Fish implicitly optimizes IDGM by construction (up to a constant factor), avoiding the computation of second-order derivative $\partial \pg/\partial \theta$.
We present this more formally for the full gradient $G$ in \Cref{theorem:fish}.

\begin{theorem}\label{theorem:fish}
Given twice-differentiable model with parameters $\theta$ and objective $l$.
Let us define the following:
\begin{align*}
    G_f &= \E[(\theta- \widetilde{\theta})] - \alpha S\cdot \bG, &\text{\small Fish update - $\alpha S\cdot$ERM grad} \\
    G_g &= - \partial \pG/\partial \theta, &\text{\small grad of $\max_{\theta}(\pG)$}
\end{align*}
where $\bG=\frac{1}{S}\sum^S_{s=1} G_s$ and is the full gradient of ERM. Then we have
\vspace{-3pt}
\begin{align*}
    \lim_{\alpha\rightarrow0}\frac{G_f \cdot G_g}{\norm{G_f}\cdot\norm{G_g}} = 1.
\end{align*}
\end{theorem}
Note that the expectation in $G_f$ is over the sampling of domains and minibatches.
\Cref{theorem:fish} indicates that when $\alpha$ is sufficiently small, if we remove the scaled \gls{ERM} gradient component $\bG$ from Fish's update, we are left with a term $G_f$ that is in similar direction to the gradient of maximizing the GIP term in IDGM, which was originally second-order.
%
%
Note that this approximation comes at the cost of losing direct control over the GIP scaling $\gamma$ ---
we therefore also derived a smoothed version of Fish that recovers this scaling term, however we find that changing the value of $\gamma$ does not make much difference empirically. See \Cref{sec:smoothfish} for more details.

The proof to \Cref{theorem:fish} can be found in \Cref{sec:taylor}. 
We follow the analysis from \citet{reptile}, which proposes Reptile for \gls{MAML}, where the relationship between inner-loop update and maximization of gradient inner product was first highlighted. 
\citet{reptile} found the GIP  term in their algorithm to be over minibatches from the \emph{same domain}, which promoted within-task generalization; in Fish we construct inner-loop using minibatches over \emph{different domains} -- it therefore instead encourages across-domain generalization.
We compare the two algorithms in further details in \Cref{sec:connection}.

We also train Fish on our simple linear dataset, with results in Table~\ref{tab:toy_results}, and see it performs similarly to IDGM -- the model assigns the most weight to the invariant feature $f_1$, and achieves $93\%$ accuracy on both train and test dataset.



\section{Experiments}

\subsection{\textsc{CdSprites-N}}
\label{sec:cdsprites}
\begin{wrapfigure}[12]{r}{0.4\textwidth}
\vspace{-12pt}
    \subcaptionbox{Train\label{fig:cdr-train}}{\includegraphics[height=2.7cm, trim={7cm 0cm 14cm 6.8cm}, clip]{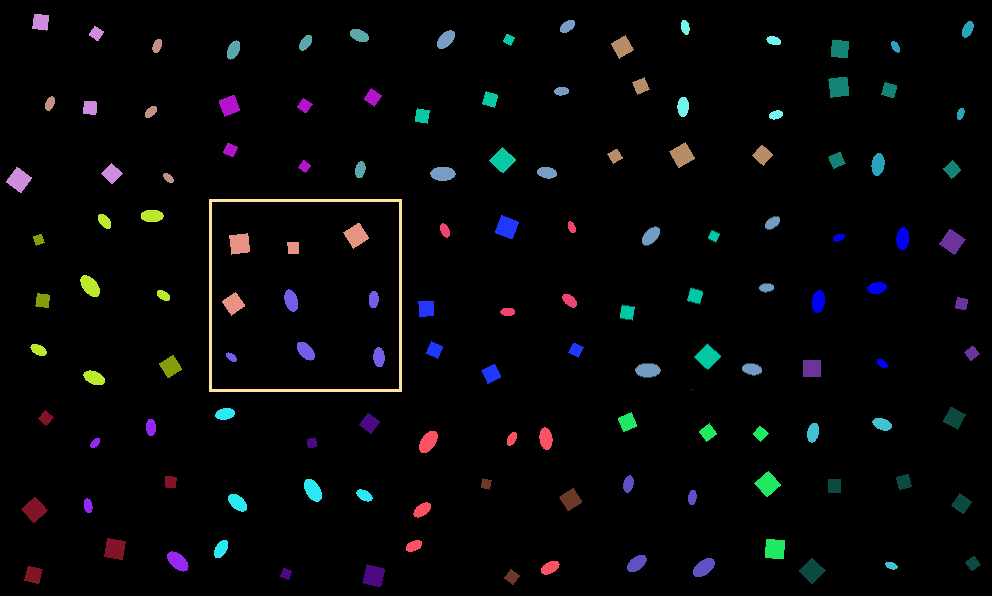}}
    \subcaptionbox{Test\label{fig:cdr-test}}{\includegraphics[height=2.7cm, trim={0cm 4.5cm 4.5cm 0cm}, clip]{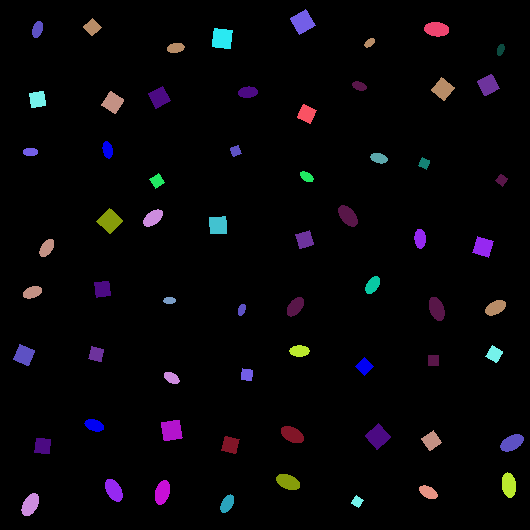}}
    \vspace*{-1ex}
    \caption{\textsc{CdSprites-N} train and test splits. Each 3x3 grid in train (e.g. yellow block) represents one domain.}\label{fig:cdr-15-sample}
\end{wrapfigure}
\textbf{Dataset}
We propose a simple shape-color dataset \textsc{CdSprites-N} based on the \textsc{dSprites} dataset \citep{dsprites}, which contains a collection of white 2D sprites of different shapes, scales, rotations and positions.
\textsc{CdSprites-N} contains $N$ domains. The goal is to classify the shape of the sprites, and there is a shape-color deterministic matching that is specific per domain.
This way we have shape as the invariant feature and color as the spurious feature.
See \Cref{fig:cdr-15-sample} for an illustration.

To construct the train split of \textsc{CdSprites-N}, we take a subset of \textsc{dSprites} that contains only 2 shapes (square and oval).
We make $N$ replicas of this subset and assign 2 colors to each, with every color corresponding to one shape (e.g. yellow block in \Cref{fig:cdr-train}, pink $\rightarrow$ squares, purple $\rightarrow$ oval).
For the test split, we create another replica of the \textsc{dSprites-N} subset, and randomly assign one of the $2N$ colors in the training set to each shape in the test set.

We design this dataset with CNN's texture bias in mind \citep{texture1, texture2}.
If the value of $N$ is small enough, the model can simply memorize the $N$ colors that correspond to each shape, and make predictions solely based on colors, resulting in poor performance on the test set where color and shape are no longer correlated.
Compared to other simple domain generalization datasets such as Digits-5 and Office-31, our dataset allows for precise control over the features that remains stable across domains and the features that change as domains change;
we can also change the number of domains $N$ easily, making it possible to examine the effect $N$ has on the performance for domain generalization.

\begin{figure}
    \centering
    \begin{subfigure}{0.32\linewidth}
        \centering
        \includegraphics[width=\linewidth, trim={0.2cm 0cm 1cm 1cm}, clip]{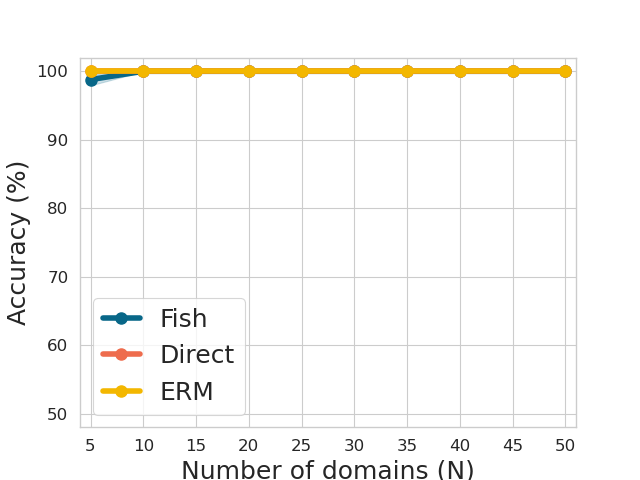}
        \caption{Train}\label{fig:cdr-results-train-}
    \end{subfigure}
    \begin{subfigure}{0.32\linewidth}
        \centering
        \includegraphics[width=\linewidth, trim={0.15cm 0cm 1cm 1cm}, clip]{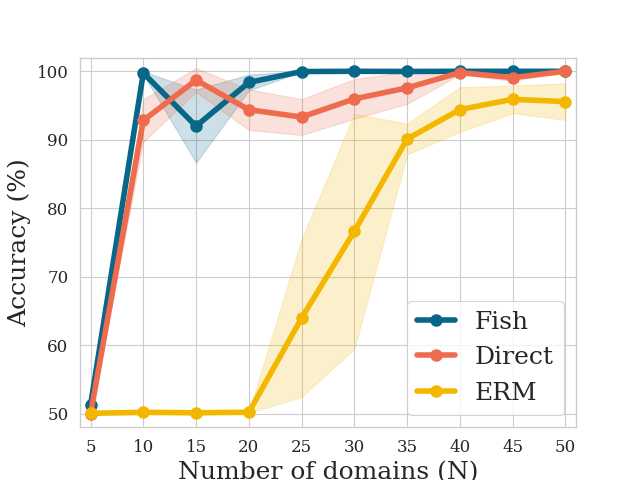}
        \caption{Test}\label{fig:cdr-results-test-}
    \end{subfigure}
    \vspace{-5pt}
    \caption{Performance on \textsc{CdSprites-N}, with $N \in [5, 50]$}\label{fig:cdr-results-}
    \vspace{-20pt}
\end{figure}
\textbf{Results}
We train the same model using three different objectives including Fish, dicrect optimization of IDGM and ERM on this dataset with number of domains $N$ ranging from $5$ to $50$.
Again, for direct optimization of IDGM, we use the normalized gradient inner product to avoid exploding gradient.

We plot the average train, test accuracy for each objective over 5 runs against the number of domains $N$ in \Cref{fig:cdr-results-}.
We can see that the train accuracy is always $100\%$ for all methods regardless of $N$ (\Cref{fig:cdr-results-train-}), while the test performance varies: \Cref{fig:cdr-results-test-} shows that {\color{c2} \textbf{direct}} optimization of IDGM (red) and {\color{c1}\textbf{Fish}} (blue) obtain the best performances, with the test accruacy rising to over $90\%$ when $N\geq 10$ and near $100\%$ when $N\geq 20$.
The predictions of {\color{c3}\textbf{ERM}} (yellow), on the other hand, remain nearly random on the test set up until $N=20$, and reach $95\%$ accuracy only for $N\geq 40$.

This experiment confirms the following:
1) the proposed IDGM objective have much stronger domain generalization capabilities compared to \gls{ERM};
2) Fish is an effective approximation of IDGM, with similar performance to its direct optimization. We also plot the gradient inner product progression of Fish vs. ERM during training in \Cref{fig:gcdsprites}, showing clearly that Fish does improve the gradient inner product across domain while ERM does not;
3) we also observe during training that Fish is about 10 times faster than directly optimizing IDGM, demonstrating its computational efficiency.

\subsection{\textsc{Wilds}}
\label{sec:wilds}
\textbf{Datasets} We evaluate our model on the \textsc{Wilds} benchmark \citep{wilds}, which contains multiple datasets that capture real-world distribution shifts across a diverse range of modalities.
We report experimental results on 6 challenging datasets in \textsc{Wilds}, and find Fish to outperform all baselines on most tasks.
A summary on the \textsc{Wilds} datasets can be found in \Cref{tab:wilds_summary}.
\begin{table*}[ht!]
\vspace{-8pt}
\centering
\caption{A summary on \textsc{Wilds} datasets. See more details on the dataset in \Cref{sec:wilds_results_app}.}\label{tab:wilds_summary}
\scalebox{0.8}{
\begin{tabular}{lllccc}
    \toprule
    \textbf{Dataset} & \textbf{Domains} & \textbf{Metric} & \textbf{Disjoint} & \textbf{Architecture}  & \textbf{\# Train examples}\\
    \midrule
    \textsc{Poverty} & 23 countries & Pearson (r) & \cmark & Resnet-18 & 10,000 \\
    \textsc{Camelyon17} & 5 hospitals & Avg. acc. & \cmark & DenseNet-121 & 302,436 \\
    \textsc{FMoW} & 16 years x 5 regions &  Avg. \& worst group acc. & \cmark & DenseNet-121 & 76,863\\
    \textsc{CivilComments} & 8 demographic groups & worst group acc. & \xmark & BERT & 269,038 \\
    \textsc{iWildCam} & 324 locations & Macro F1 & \cmark & ResNet-50 & 142,202 \\
    \textsc{Amazon} & 7,676 reviewers & 10th percentile acc. & \cmark & BERT &1,000,124  \\
    \bottomrule
\end{tabular}}
\end{table*}
For hyperparameters including learning rate, batch size, choice of optimizer and model architecutre, we follow the exact configuration as reported in the \textsc{Wilds} benchmark.
Importantly, we also use the same model selection strategy used in \textsc{Wilds} to ensure a fair comparison.
See details in \Cref{sec:hyper_app}.

\begin{table}[H]
\centering
\vspace{-10pt}
\caption{Results on \textsc{Wilds} benchmark.}
\scalebox{0.7}{
\begin{tabular}{lccccccccccc}    \toprule
     & {
    \textsc{PovertyMap}} & \hphantom{a} &{
    \textsc{Camelyon17}} & \hphantom{a} & {
    \textsc{FMoW}} & \hphantom{a} & {
    \textsc{CivilComments}} & \hphantom{a} & {
    \textsc{iWildCam}} &  \hphantom{a} & {
    \textsc{Amazon}} \\
    \cmidrule{2-2} \cmidrule{4-4} \cmidrule{6-6} \cmidrule{8-8} \cmidrule{10-10} \cmidrule{12-12}
    & Pearson r & & Avg. acc. (\%) & & Worst acc. (\%) & & Worst acc. (\%) & & Macro F1 & &10-th per. acc. (\%) \\
    \midrule
    \textbf{Fish} & \textbf{0.80 \std{1e-2}} & & \textbf{74.7} \std{7e-2} & & \textbf{34.6 \std{0.00}} & & \textbf{72.8 \std{0.0}} & & 22.0 \std{0.0}            & & \textbf{53.3 \std{0.0}} \\
    IRM           & 0.78 \std{3e-2}          & & 64.2 \std{8.1}         & & 33.5 \std{1.35}          & & 66.3 \std{2.1}            & & 15.1 \std{4.9}            & & 52.4 \std{0.8}\\
    Coral         & 0.77 \std{5e-2}          & & 59.5 \std{7.7}         & & 31.0 \std{0.35}          & & 65.6 \std{1.3}            & & \textbf{32.8\std{0.1}}    & & 52.9 \std{0.8} \\
    Reweighted    & -                        & &  -                     & & -                        & & 66.2 \std{1.2}            & & -                         & & 52.4 \std{0.8}\\
    GroupDRO      & 0.78 \std{5e-2}          & & 68.4 \std{7.3}         & & 31.4 \std{2.10}           & & 69.1 \std{1.8}            & & 23.9 \std{2.1}            & & 53.5 \std{0.0} \\
    ERM           & 0.78 \std{3e-2}          & & 70.3 \std{6.4}         & & 32.8 \std{0.45}          & & 56.0 \std{3.6}            & & 31.0 \std{1.3}   & & \textbf{53.8 \std{0.8}}\\
    ERM (ours)    & 0.77 \std{5e-2}          & & 70.5 \std{12.1}        & & 30.9 \std{1.53}          & & 58.1 \std{1.7}            & & 25.1 \std{0.2}   & & \textbf{53.3 \std{0.8}}\\
    \bottomrule
\end{tabular}\label{tab:wilds_results}}
\end{table}

\textbf{Results}
See a summary of results in \Cref{tab:wilds_results}, where we use the metrics recommended in \textsc{Wilds} for each dataset.
Again, following practices in \textsc{Wilds}, all results are reported over 3 random seed runs, apart from \textsc{Camelyon17} which is reported over 10 random seed runs. 
We included additional results as well as a in-depth discussion on each dataset in \Cref{sec:wilds_results_app}.
We make the following observations:
\begin{enumerate}[leftmargin=0.4cm]
    \item \textbf{Strong performance across datasets:} Considering results on all 6 datasets, Fish is the best performing algorithm on \textsc{Wilds}. It outperforms all baseline on 4 datasets and achieves similar level of performance to the best method on the other 2 (\textsc{Amazon} and \textsc{iWildCam}). Fish's strong performance on different types of data and architectures such as \textsc{ResNet} \citep{resnet}, \textsc{DenseNet} \citep{dense} and \textsc{BERT} \citep{bert} demonstrated it's capability to generalize to a diverse variety of tasks;
    \item \textbf{Strong performance on different domain generalization tasks:} We make special note the \textsc{CivilComments} dataset captures \emph{subpopulation shift} problems, where the domains in test are a subpopulation of the domains in train, while all other \textsc{Wilds} datasets depicts \emph{pure domain generalization} problems, where the domains in train and test are disjointed. As a result, the baseline models for \textsc{CivilComments} selected by the \textsc{Wilds} benchmark are different from the methods used in all other datasets, and are tailored to avoiding systematic failure on data from minority subpopulations. We see that Fish works well in this setting too without any changes or special sampling strategies (used for baselines on \textsc{CivilComments}, see more in \Cref{tab:civil_results}), demonstrating it's capability to perform in different domain generalization scenarios;
    \item \textbf{Failure mode of domain generalization algorithms:} We noticed that on \textsc{iWildCam} and \textsc{Amazon}, ERM is the best algorithm, outperforming all domain generalization algorithms except for Fish on \textsc{Amazon}.  We believe that these domain generalization algorithms failed due to the large number of domains in these two datasets --- 324 for \textsc{iWildCam} and 7,676 for \textsc{Amazon}. This is a common drawback of current domain generalization literature and is a direction worth exploring.
\end{enumerate}

\subsection{\textsc{DomainBed}}
\begin{table}
\centering
   \caption{Test accuracy ($\%$) on \textsc{DomainBed} benchmark.}
    \scalebox{0.7}{
    \begin{tabular}{lccccccccccc}
    \toprule
       & ERM            & IRM            & GroupDRO       & Mixup          & MLDG           & Coral          & MMD            & DANN           & CDANN          & Fish (ours)         \\ \midrule
      CMNIST    & 52.0 \std{0.1} & 51.8 \std{0.1} & 52.0 \std{0.1} & 51.9 \std{0.1} & 51.6 \std{0.1} & 51.7 \std{0.1} & 51.8 \std{0.1} & 51.5 \std{0.3} & 51.9 \std{0.1} & 51.6 \std{0.1}\\
      RMNIST    & 98.0 \std{0.0} & 97.9 \std{0.0} & 98.1 \std{0.0} & 98.1 \std{0.0} & 98.0 \std{0.0} & 98.1 \std{0.1} & 98.1 \std{0.0} & 97.9 \std{0.1} & 98.0 \std{0.0} & 98.0 \std{0.0} \\
      VLCS      & 77.4 \std{0.3} & 78.1 \std{0.0} & 77.2 \std{0.6} & 77.7 \std{0.4} & 77.1 \std{0.4} & 77.7 \std{0.5} & 76.7 \std{0.9} & 78.7 \std{0.3} & 78.2 \std{0.4} & 77.8 \std{0.3}\\
      PACS      & 85.7 \std{0.5} & 84.4 \std{1.1} & 84.1 \std{0.4} & 84.3 \std{0.5} & 84.8 \std{0.6} & 86.0 \std{0.2} & 85.0 \std{0.2} & 84.6 \std{1.1} & 82.8 \std{1.5} & 85.5 \std{0.3} \\
      OfficeHome& 67.5 \std{0.5} & 66.6 \std{1.0} & 66.9 \std{0.3} & 69.0 \std{0.1} & 68.2 \std{0.1} & 68.6 \std{0.4} & 67.7 \std{0.1} & 65.4 \std{0.6} & 65.6 \std{0.5} & 68.6 \std{0.4} \\
      TerraInc  & 47.2 \std{0.4} & 47.9 \std{0.7} & 47.0 \std{0.3} & 48.9 \std{0.8} & 46.1 \std{0.8} & 46.4 \std{0.8} & 49.3 \std{1.4} & 48.7 \std{0.5} & 47.6 \std{0.8} & 45.1 \std{1.3} \\
      DomainNet & 41.2 \std{0.2} & 35.7 \std{1.9} & 33.7 \std{0.2} & 39.6 \std{0.1} & 41.8 \std{0.4} & 41.8 \std{0.2} & 39.4 \std{0.8} & 38.4 \std{0.0} & 38.9 \std{0.1} & 42.7 \std{0.2} \\\midrule
      Average   & 67.0           & 66.0           & 65.5           & 67.1           & 66.8           & 67.2                    & 66.8           & 66.4           & 66.1           & 67.1          \\
    \bottomrule
    \end{tabular}
    }\label{tab:domainbed}
\end{table}

\textbf{Datasets} While \textsc{Wilds} is a challenging benchmark capturing realistic distribution shift, to test our model under the synthetic-to-real transfer setting and provide more comparisons to SOTA methods, we also performed experiments on the \textsc{DomainBed} benchmark \citep{deepcoral}. 

\textsc{DomainBed} is a testbed for domain generalization that implements consistent experimental protocols across SOTA methods to ensure fair comparison.
It contains 7 popular domain generalization datasets, including Colored MNIST \citep{irm}, Rotated MNIST \citep{ghifary2015domain}, VLCS \citep{vlcs}, PACS \citep{pacs}, OfficeHome \citep{officehome}, Terra Incognita \citep{beery2018recognition} and DomainNet \citep{peng2019moment}, and
offers comparison to a variety of SOTA domain generalization methods, including IRM \citep{irm}, Group DRO \citep{groupdro0, groupdro1}, Mixup \citep{yan2020improve}, MLDG \citep{li2018learning}, Coral \citep{sun2016deep}, MMD \citep{li2018domain}, DANN \citep{ganin2016domain} and CDANN \citep{li2018deep}.

\textbf{Results} Following recommendations in \textsc{DomainBed}, we report results using training domain as validation set for model selection.
See results in \Cref{tab:domainbed}, reported over 5 random trials. 
Averaging the performance over all 7 datasets, Fish ranks second out of 10 domain generalization methods. 
It performs only marginally worse than Coral (0.1$\%$), and is one of the three methods that performs better than ERM.
This showcases Fish's effectiveness on domain generalization datasets with stronger focus to synthetic-to-real transfer, which again demonstrates its versatility and robustness on different domain generalization tasks.

\subsection{Analysis}
\label{sec:track}

\begin{wrapfigure}[12]{r}{0.35\textwidth}
\vspace{-20pt}
\includegraphics[width=\linewidth]{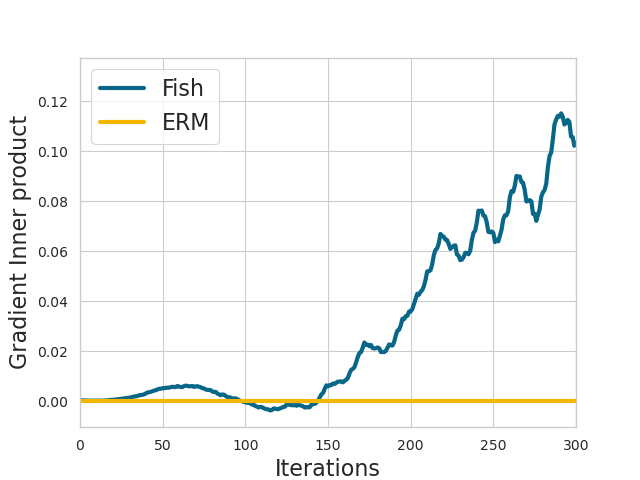}
\caption{Gradient inner product values during the training for \textsc{CdSprites-N} (N=15).}\label{fig:gradprog}
\end{wrapfigure}

\textbf{Tracking gradient inner product}
In \Cref{fig:gradprog}, we demonstrate the progression of inter-domain gradient inner products during training using different objectives.
We train both {\color{c1}\textbf{Fish}} (blue) and {\color{c3}\textbf{ERM}} (yellow) untill convergence while tracking the normalized gradient inner products between minibatches from different domains used in each inner-loop.
To ensure a fair comparison, we use the exact same sequence of data for Fish and ERM (see \Cref{sec:trackinggradipm} for more details).

From \Cref{fig:gradprog}, it is clear that during training, the normalized gradient inner product of Fish increases, while that for ERM stays at the same value.
The observations here shows that Fish is indeed effective in increasing/maintaining the level of inter-domain gradient inner product.

We conduct the same gradient inner product tracking experiments for the \textsc{Wilds} datasets we studied as well to shed some lights on its efficiency --- see \Cref{sec:track_grad_app} for results.

\textbf{Random Grouping}
 We conducted experiments where data are grouped randomly instead of by domain for the inner-loop update.
By doing so, we are still maximizing the inner product between minibatches, however it no longer holds that each minibatch contain data from one domain only.
We therefore expect the results to be slightly worse than Fish, and the bigger the domain gap is, the more advantage Fish has against the random grouping strategy.
We show the results for random grouping (\textit{Fish, RG}) in \Cref{tab:rg}. 

\begin{table}[H]
\centering
    \caption{Ablation study on random grouping: test accuracy on different datasets.}\label{tab:rg}
    \scalebox{0.8}{
    \begin{tabular}{lccccc}
    \toprule
    & \textsc{cDsprites}(N=10) & \textsc{FMoW} & \textsc{VLCS} & \textsc{PACS} & OfficeHome \\\midrule
    Fish      & 100.0 \std{0.0} & 34.3 \std{0.6} & 77.6 \std{0.5} & 85.5 \std{0.3} & 68.6 \std{0.9} \\
    Fish, RG  & 50.0 \std{0.0}& 33.4 \std{1.7} & 77.7 \std{0.3} & 83.9 \std{0.7} & 66.5 \std{1.0}  \\
    ERM       & 50.0 \std{0.0}& 31.7 \std{1.0} & 77.5 \std{0.4} & 85.5 \std{0.2}  & 66.5 \std{0.3}  \\
    \bottomrule
    \end{tabular}}
\end{table}
As expected, the random grouping strategy performs worse than Fish on all datasets.
This is the most prominent on \textsc{cDsprites} with 10 domains (N=10), where Fish achieves $100\%$ test accuracy and both random grouping and ERM predicts randomly on the test split.
The experiment demonstrated that the effectiveness of our algorithm largely benefited from the domain grouping strategy, and that maximising the gradient inner product between random batches of data does not achieve the same domain generalization performance.

\textbf{Ablation studies on hyper-parameters}
We provide ablation studies on the learning rate of Fish's inner-loop $\alpha$, meta step $\epsilon$, and number of inner loop steps $N$ in \Cref{sec:abla_app}.
We also study the effect of fine-tuning on pretrained models at different stage of convergence in \Cref{sec:pretrained}.


\section{Conclusion}

In this paper we presented inter-domain gradient matching (IDGM) for domain generalization.
To avoid costly second-order computations, we approximated IDGM with a simple first-order algorithm, Fish.
We demonstrated our algorithm's capability to learn from invariant features (as well as ERM's failure to do so) using simple datasets such as \textsc{CdSprites-N} and the linear example.
We then evaluated the model's performance on \textsc{Wilds} and \textsc{DomainBed}, demonstrating that Fish performs well on different subgenres of domain generalization, and surpasses baseline performance on a diverse range of vision and language tasks using different architectures such as DenseNet, ResNet-50 and BERT.
Our experiments can be replicated with 1500 GPU hours on NVIDIA V100.

Despite its strong performance, similar to previous work on domain generalization, when the number of domains is large Fish struggles to outperform ERM.
We are currently investigating approaches by which Fish can be made to scale to datasets with orders of magnitude more domains and expect to report on this improvement in our future work.



\FloatBarrier
\clearpage



\bibliography{main}

\clearpage
\appendix

\section{Taylor Expansion of Reptile and Fish Inner-loop Update}\label{sec:taylor}
In this section we provide proof to \Cref{theorem:fish}.
We reproduce and adapt the proof from \cite{reptile} in the context of Fish, for completeness.

We demonstrate that when the inner-loop learning rate $\alpha$ is small, the direction of $G_f$ aligns with that of $G_g$, where 
\begin{align}
    G_f &= \E\left[(\theta- \tilde{\theta})\right] - \alpha S\cdot \bG, \label{eq:gf} \\
    G_g &= - \partial \pG/\partial \theta, \label{eq:gd}
\end{align}
%

\paragraph{Expanding $G_g$}
$G_g$ is the gradient of maximizing the gradient inner product (GIP).
\begin{align}
    G_g = - \frac{2}{S(S-1)} \sum_{i,j\in S}^{i\neq j} \frac{\partial}{\partial\theta} G_i\cdot G_j \label{eq:finalgd}
\end{align}
\paragraph{Expanding $G_f$}
To write out $G_f$, we need to derive the gradient update of Fish, $\theta-\tilde{\theta}$.
Let us first define some notations.

For each inner-loop with $S$ steps of gradient updates, we assume a loss functions $l$ as well as a sequence of inputs $\{\d_i\}_{i=1}^S$, where $\d_i:=\{x_b, y_b\}_{b=1}^B$ denotes a minibatch at step $i$ randomly drawn from one of the available domains in $\{\D_1,\cdots, \D_S\}$.
For reasons that will become clearer later, take extra note that the subscript $i$ here denotes the index of step, rather than the index of domain.
We also define the following:
\begin{align}
    &\tg_i = \E_{\d_i} \left[\frac{\partial l((x,y); \theta_i)}{\partial \theta_i} \right] &\text{(gradient at step $i$, wrt $\theta_i$)}\\
    &\theta_{i+1} = \theta_i - \alpha \tg_i  &\text{(sequence of parameters)}\\
    & \g_i = \E_{\d_i} \left[\frac{\partial l((x,y); \theta_1)}{\partial \theta_1} \right] &\text{(gradient at step $i$, wrt $\theta_1$)} \label{eq:gi}\\
    & H_i = \E_{\d_i} \left[\frac{\partial^2 l((x,y); \theta_1)}{\partial \theta_1^2} \right] & \text{(Hessian at initial point)}    
\end{align}
In the following analysis we omit the expectation $\E_{\d_i}$ and input $(x,y)$ to $l$ and instead denote the loss at step $i$ as $l_i$.
Performing second-order Taylor approximation to $\tg_i$ yields:
\begin{align}
    \tg_i &= l_i'(\theta_i) \\ 
    &= l_i'(\theta_1) + l_i''(\theta_1)(\theta_i - \theta_1) + \underbrace{\mathcal{O}(\lVert \theta_i - \theta_1 \rVert^2)}_{=\mathcal{O}(\alpha^2)}\\
    &= g_i + H_i(\theta_i - \theta_1) + \mathcal{O}(\alpha^2)\\
    &=\g_i - \alpha H_i \sum^{i-1}_{j=1} \tg_j + \mathcal{O}(\alpha^2). \label{eq:gi-so-}
\end{align}
Applying first-order Taylor approximation to $\tg_j$ gives us
\begin{align}
    \tg_j = g_j + \mathcal{O}(\alpha), \label{eq:gj-fo}
\end{align}
plugging this back to \Cref{eq:gi-so-} yields:
\begin{align}
    \tg_i &= g_i - \alpha H_i \sum^{i-1}_{j=1} g_j + \mathcal{O}(\alpha^2). \label{eq:gi-so}
\end{align}
For simplicity reason, let us consider performing two steps in inner-loop updates, i.e. $S=2$.
We can then write the gradient of Fish $\theta-\tilde{\theta}$ as
\begin{align}
    \theta-\tilde{\theta} &= \alpha(\tg_1 + \tg_2) \\
    &= \alpha(\underbrace{g_1 + g_2}_{\scriptsize\circled{1}}) - \alpha^2 \underbrace{H_2g_1}_{\scriptsize\circled{2}} + \mathcal{O}(\alpha^3)\label{eq:rf-grad}.
\end{align}
%
%
%
%
%
Furthermore, taking the expectation of $\theta-\tilde{\theta}$ under minibatch sampling gives us
\begin{align*}
    \circled{1} &= \E_{1,2}\left[g_1+g_2\right] = G_1 + G_2\\ 
    \circled{2} &= \E_{1,2}\left[{H}_2g_1\right]= \E_{1,2}\left[{H}_1g_2\right] &\text{(interchanging indices)}\\
    &= \frac{1}{2} \E_{1,2}\left[ {H}_2g_1 + {H}_1g_2\right] &\text{(averaging last two eqs)}\\
    &=  \frac{1}{2} \E_{1,2}\left[ \frac{\partial(\g_1 \cdot \g_2)}{\partial \theta_1}\right]\\
    &= \frac{1}{2}\cdot \frac{\partial(G_1\cdot G_2)}{\partial \theta_1}  \label{eq:gradprod}
\end{align*}
Note that the only reason we can interchange the indices in {\scriptsize \circled{2}} is because the subscripts represent steps in the inner loop rather than index of domains.
Plugging {\scriptsize \circled{1}}, {\scriptsize \circled{2}} in \Cref{eq:rf-grad} yields:
\begin{align}
    \E[\theta-\tilde{\theta}]=\alpha(G_1+G_2) + \frac{\alpha^2}{2}\cdot \frac{\partial(G_1\cdot G_2)}{\partial \theta_1} + \mathcal{O}(\alpha^3)
\end{align}

%
%

We can also expand this to the general case where $S\geq 2$:
\begin{align}
    &\E[\theta-\tilde{\theta}] \notag \\
    & \quad = \alpha \sum_{s=1}^S G_s  - \frac{\alpha^2}{S(S-1)}\sum_{i,j \in S}^{i\neq j} \frac{\partial(G_i \cdot G_j)}{\partial\theta_1}  + \mathcal{O}(\alpha^3).\label{eq:fish_update}
\end{align}

The second term in \Cref{eq:gf} is $\bG$, which is the full gradient of ERM defined as follow:
\begin{align}
    \bG=\frac{1}{S} \sum^S_{s=1}G_s.\label{eq:erm}
\end{align}
Plugging \Cref{eq:fish_update} and \Cref{eq:erm} to \Cref{eq:gf} yields
\begin{align}
    G_f &= \E[\theta-\tilde{\theta}] - \alpha S \bG \\
    &=- \frac{\alpha^2}{S(S-1)}\sum_{i,j \in S}^{i\neq j} \frac{\partial}{\partial\theta_1} G_i \cdot G_j 
    \label{eq:finalgf}
\end{align}
Comparing \Cref{eq:finalgd} to \Cref{eq:finalgf}, we have:
\begin{align*}
    \lim_{\alpha\rightarrow0}\frac{G_f \cdot G_g}{\norm{G_f} \norm{G_g}} = 1.
\end{align*}

\subsection{Fish and Reptile: Differences and Connections} \label{sec:connection}
As we introduced, our algorithm Fish is inspired by Reptile, a \gls{MAML} algorithm.

Even though meta learning and domain generalization both study $N$-way, $K$-shot problems, there are some distinct differences that set them apart.
The most prominent one is that in meta learning, some examples in the test dataset will be made available at test time ($K>0$), while in domain generalization no example in the test dataset is seen by the model ($K=0$);
another important difference is that while domain generalization aims to train models that perform well on an unseen distribution of the \emph{same task}, meta-learning assumes \emph{multiple tasks} and requires the model to quickly learn an unseen task using only $K$ examples.

Due to these differences, it does not make sense in general to use \gls{MAML} framework in domain generalization. 
As it turns out however, the idea of aligning gradients to improve generalization is relevant to both methods ---
The fundamental difference here that \gls{MAML} algorithms such as Reptile aligns the gradients between batches from the \emph{same} task \cite{reptile}, while Fish aligns those between batches from \emph{different} tasks.

To see how this is ahiceved, let us have a look at the algorithmic comparison between {\color{c1}\textbf{Fish}} (blue) and {\color{c4}\textbf{Reptile}} (green) in \Cref{alg:reptile}.
As we can see, the key difference between the algorithm of Fish and Reptile is that Reptile performs its inner-loop using minibatches from the \emph{same} task, while Fish uses minibatches from \emph{different} tasks (\textit{l4-8}).
Based on the analysis in \citet{reptile} (which we reproduce in \Cref{sec:taylor}), this is why Reptile maximizes the within-task gradient inner products and Fish maximizes the across-task gradient inner products.
\begin{algorithm}[t]
    \begin{algorithmic}[1] 
    \FOR{i = $1, 2, \cdots$}
        \STATE $\tilde{\theta} \gets \theta$
        \STATE {\color{c4} Sample task $\D_t \sim \{\D_1,\cdots,\D_T\}$}
        \FOR{${\color{c4}s \in \{1, \cdots, S\}}$ \textbf{or} ${\color{c1}\D_t \in \{\D_1, \cdots, \D_T\}}$}
            \STATE Sample batch $\bm{d}_t \sim \D_t$
            \STATE $\g_t = \partial \L(\bm{d}_t;\tilde{\theta})/\partial\tilde{\theta}$
            \STATE Update $\tilde{\theta} \gets \tilde{\theta} - \alpha\g_t$
        \ENDFOR
        \STATE Update $\theta \gets \theta + \displaystyle \epsilon(\tilde{\theta}-\theta)$
    \ENDFOR
    \end{algorithmic}
\caption{Black fonts denote steps used in \textbf{both algorithms},  colored fonts are steps unique to {\color{c1}\textbf{Fish}} or {\color{c4}\textbf{Reptile}}.}\label{alg:reptile}
\end{algorithm}

\begin{figure}
    \centering
    \begin{subfigure}{0.4\linewidth}
        \centering
        \includegraphics[width=\linewidth, trim={0.6cm 0.2cm 1cm 1cm}, clip]{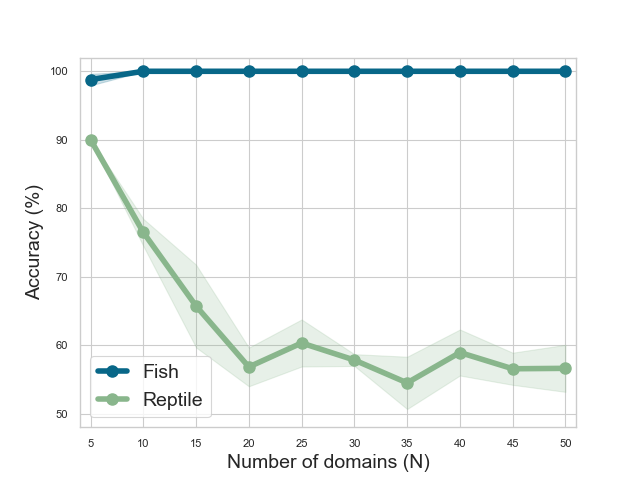}
        \caption{Train}\label{fig:cdr-results-train}
    \end{subfigure}
    \begin{subfigure}{0.4\linewidth}
        \centering
        \includegraphics[width=\linewidth, trim={0.5cm 0.2cm 1cm 1cm}, clip]{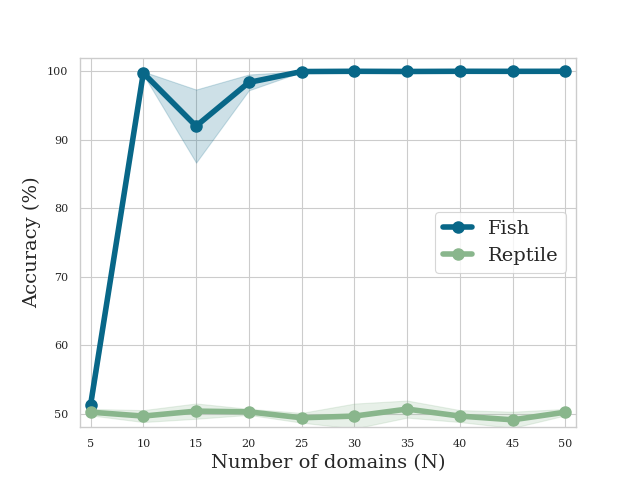}
        \caption{Test}\label{fig:cdr-results-test}
    \end{subfigure}
    \caption{Performance on \textsc{CdSprites-N}, with $N \in [5, 50]$}\label{fig:cdr-results-rf}
\end{figure}

A natural question to ask here is -- how does this affect their empirical performance?
In \Cref{fig:cdr-results-rf}, we show the train and test performance of {\color{c1}\textbf{Fish}} (blue) and {\color{c4}\textbf{Reptile}} (green) on \textsc{CdSprites-N}.
We can see that despite the algorithmic similarity between Fish and Reptile, the two methods behave very differently on this domain generalization task:
while Fish's test accuracy goes to $100\%$ at $N =10$, Reptile's test performance is always $50\%$ regardless of $N$.
Moreover, we observe a dip in Reptile's training performance early on, with the accuracy plateaus at $56\%$ when $N>20$.
%
Reptile's poor performance on this dataset is to be expected since its inner-loop is designed to encourage within-domain generalization, which is not helpful for learning what's invariant across domains.

\section{SmoothFish: a more general algorithm}\label{sec:smoothfish}
\subsection{Derivation}
We conclude in \Cref{sec:taylor} that a component of Fish's update $G_f =  \E[\theta-\tilde{\theta}] - \alpha S\cdot \bar{G}$ is in the same direction as the gradient of GIP, $G_g$.
It is therefore possible to have explicit control over the scaling of the GIP component in Fish, similar to the original IDGM objective, by writing the following:
\begin{align}
    G_{\text{sm}} = \alpha S \cdot \bar{G} + \gamma \left(\E[\theta - \tilde{\theta}]-\alpha S \cdot \bar{G} \right).
\end{align}
By introducing the scaling term $\gamma$, we have better control on how much the objective focus on inner product vs average gradient.
Note that $\gamma=1$ recovers the original Fish gradient, and when $\gamma=0$ the gradient $G_{\text{sm}}$ is equivalent to ERM's gradient with learning rate $\alpha S$.
We name the resulting algorithm SmoothFish. See \Cref{alg:smoothfish}.
\setlength{\textfloatsep}{1.5ex}
\begin{algorithm}[t]
    \begin{algorithmic}[1]
    \makeatletter
    \newcommand{\ourlinenumber}[1]{
    \let\old@ALC@lno=\ALC@lno
    \renewcommand{\ALC@lno}{\raggedright\begin{small}#1\end{small}
    \let\ALC@lno=\old@ALC@lno}
    }
    \makeatother
    \FOR{iterations = $1, 2, \cdots$}
        \STATE ${\widetilde{\theta}} \gets \theta$
        \FOR{$\D_i \in \texttt{permute}(\{\D_1, \D_2, \cdots, \D_S\})$}
            \STATE Sample batch  ${d}_i \sim \D_i$
            \STATE $\displaystyle g_i = \E_{\d_i}\left[\frac{\partial l((x,y);{\theta})}{\partial{{\theta}}}\right]$\COMMENT{Grad wrt $\theta$}
            \STATE $\displaystyle \tg_i = \E_{\d_i}\left[\frac{\partial l((x,y);\widetilde{\theta})}{\partial{\widetilde{\theta}}}\right]$\COMMENT{Grad wrt $\widetilde{\theta}$}
            \STATE Update ${\widetilde{\theta}} \gets \widetilde{\theta} - \alpha \tg_i$
        \ENDFOR
        \STATE $\bg =\displaystyle \frac{1}{S} \sum_{s=1}^S g_i,\ g_{\text{sm}} = \alpha S\bg + \gamma\left( (\widetilde{\theta}-\theta) - \alpha S \bg\right)$
        \STATE Update $\theta \gets \theta + \displaystyle \epsilon g_{\text{sm}}$
    \ENDFOR
    \end{algorithmic}
\caption{Smoothed version of Fish, which allows to get approximate gradients for the general form of \Cref{eq:obj_ip}.}
\label{alg:smoothfish}
\end{algorithm}
\begin{figure}[H]
\centering
\begin{subfigure}{0.32\linewidth}
    \centering
    \includegraphics[width=\linewidth]{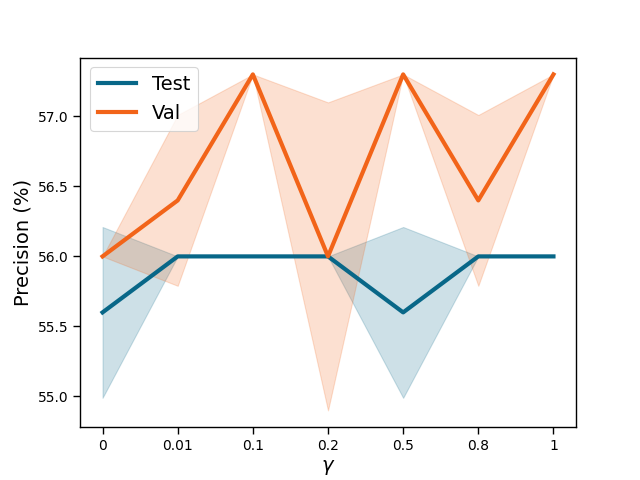}
    \caption{\textsc{Amazon}} \label{fig:samazon}
\end{subfigure}
\begin{subfigure}{0.32\linewidth}
    \centering
    \includegraphics[width=\linewidth]{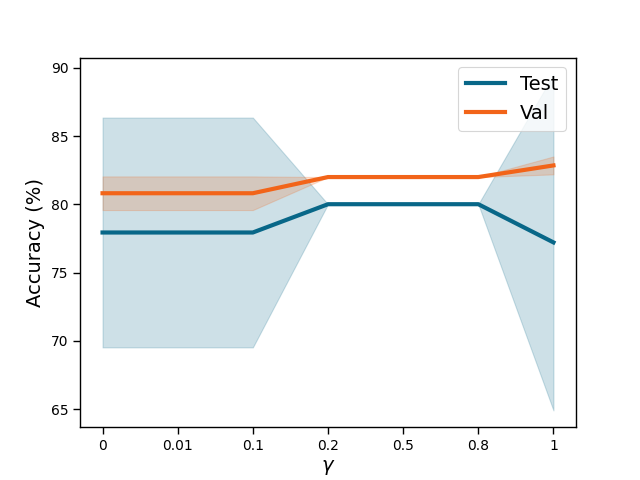}
    \caption{\textsc{Camelyon17}} \label{fig:scam}
\end{subfigure}
\begin{subfigure}{0.32\linewidth}
    \centering
    \includegraphics[width=\linewidth]{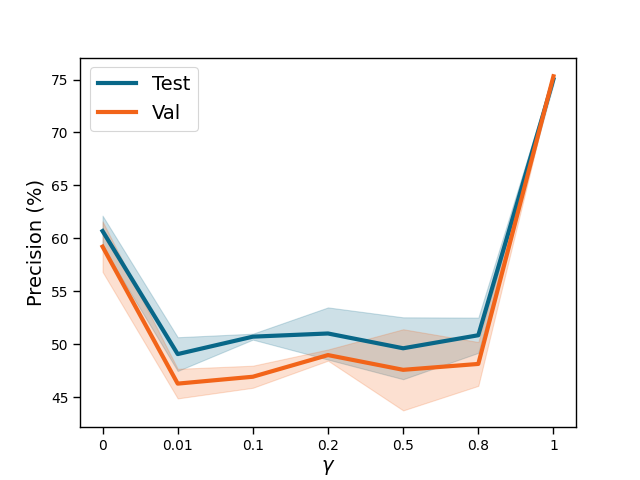}
    \caption{\textsc{CivilComments}} \label{fig:scivil}
\end{subfigure}
\begin{subfigure}{0.32\linewidth}
    \centering
    \includegraphics[width=\linewidth]{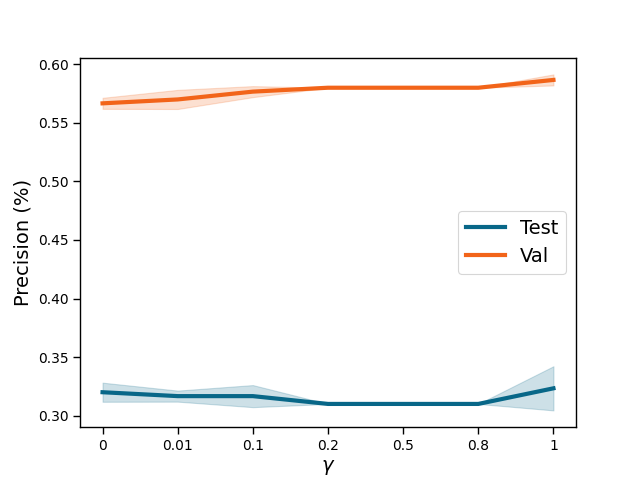}
    \caption{\textsc{FMoW}} \label{fig:sfmow}
\end{subfigure}
\begin{subfigure}{0.32\linewidth}
    \centering
    \includegraphics[width=\linewidth]{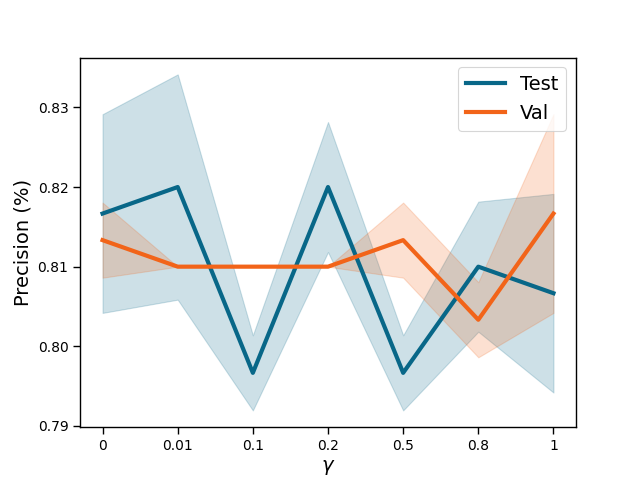}
    \caption{\textsc{Poverty}} \label{fig:spoverty}
\end{subfigure}
\begin{subfigure}{0.32\linewidth}
    \centering
    \includegraphics[width=\linewidth]{images/smoothfish/poverty.png}
    \caption{\textsc{iWildCam}} \label{fig:siwc}
\end{subfigure}
\caption{Results on \textsc{Wilds} using SmoothFish with $\gamma$ ranging from 0 to 1.} \label{fig:app_smooth_result}
\end{figure}
\subsection{Results}
We run experiments on the 6 datasets in \textsc{Wilds} using SmoothFish, with $\gamma$ ranging in $[0.1, 0.2, 0.5, 0.8]$. 
We also include results for $\gamma=0$ (equivalent to ERM) and $\gamma=1$ (equivalent to Fish).
See results in \Cref{fig:app_smooth_result}.
The other hyperparameters including $\alpha$, meta steps, $\epsilon$ used here are the same as the ones used in our main experiments section.

\section{Discussions and Results on \textsc{Wilds}}\label{sec:wilds_results_app}
We provide a more detailed summary of each dataset in \Cref{tab:wilds_deets}.
Some entries in $\texttt{$\#$ Domains}$ are omitted because the domains in each split overlap.
Note that in this paper we report the results on \textsc{Wilds} v1 --- the benchmark has been updated since with slightly different dataset splits.
We are currently working on updating our results to v2 of \textsc{Wilds}.

\begin{table*}[ht!]
\centering
\caption{Details of the 6 \textsc{Wilds} datasets we experimented on.}\label{tab:wilds_deets}
\resizebox{\textwidth}{!}{
\begin{tabular}{llllcccccccc}
    \toprule
    \multirow{2}{*}{Dataset} & \multirow{2}{*}{Domains ($\#$ domains)} & \multirow{2}{*}{Data ($x$)} & \multirow{2}{*}{Target ($y$)} & \multicolumn{3}{c}{$\#$ Examples} & & \multicolumn{3}{c}{$\#$ Domains}\\
    \cmidrule{5-7}\cmidrule{9-11}
    &&&&train&val&test&&train&val&test \\
    \midrule
    \textsc{FMoW} & Time (16), Regions (5) & Satellite images & Land use (62 classes) & 76,863 & 19,915 & 22,108 & & 11, - & 3, - & 2, - \\
    \textsc{Poverty} & Countries (23), Urban/rural (2) & Satellite images & Asset (real valued) & 10,000 & 4,000 &  4,000 & & 13, - & 5, - & 5, - \\
    \textsc{Camelyon17} & Hospitals (5) & Tissue slides & Tumor (2 classes) & 302,436 & 34,904 & 85,054 & & 3 & 1 & 1 \\
    \textsc{CivilComments} & Demographics (8) & Online comments & Toxicity (2 classes) & 269,038 & 45,180 & 133,782  & & - & - & - \\
    \textsc{iWildCam2020} & Trap locations (324) & Photos & Animal species (186 classes) & 142,202 & 20,784 & 38,943 & & 245 & 32 & 47 \\ 
    \textsc{Amazon} & Reviewers (7,676) & Product reviews & Star rating (5 classes) & 1,000,124 & 100,050 & 100,050 & & 5,008 & 1,334 & 1,334 \\ 
    \bottomrule
\end{tabular}}
\end{table*}

\begin{center}
\begin{minipage}[t]{0.48\linewidth}
\vspace{-30pt}
\begin{table}[H]
\centering
\caption{Results on \textsc{PovertyMap-Wilds}. }
\scalebox{0.8}{
\begin{tabular}{lcc}
    \toprule
    Method & Val. Pearson r & \textbf{Test Pearson r} \\
    \midrule
    \textbf{Fish} & 0.81 \std{6e-3} & {\textbf{0.81 \std{9e-3}}}\\
    IRM & 0.81 \std{4e-2} & 0.78 \std{3e-2}\\
    ERM & 0.80 \std{3e-2} & 0.78 \std{3e-2}\\
    ERM (ours) & 0.80 \std{3e-2} & 0.77 \std{5e-2}\\
    Coral & 0.80 \std{4e-2} & 0.77 \std{5e-2}\\
    \bottomrule
\end{tabular}\label{tab:poverty_results}}
\end{table}
\end{minipage}\begin{minipage}[t]{0.48\linewidth}
\vspace{-30pt}
\begin{table}[H]
\centering
\caption{Results on \textsc{Camelyon17-Wilds}.}
\scalebox{0.8}{
\begin{tabular}{lcc}
    \toprule
    Method & Val. Accuracy ($\%$) &\textbf{ Test Accuracy ($\%$)} \\
    \midrule
    \textbf{Fish} & 82.5 \std{1.2} & \textbf{79.5} \std{6.0}  \\
    \gls{ERM} & 84.3 \std{2.1} & 73.3 \std{9.9} \\
    \gls{ERM} (ours) & 84.1 \std{2.4} & 70.5 \std{12.1} \\
    IRM &  86.2 \std{2.1} & 60.9 \std{15.3} \\
    Coral & 86.3 \std{2.2} & 59.2 \std{15.1} \\
    \bottomrule
\end{tabular}\label{tab:camelyon_results}}
\end{table}
\end{minipage}
\end{center}

\subsection{\textsc{PovertyMap-Wilds}}
\label{sec:poverty-wilds}
\begin{center}
    \emph{Task: Asset index prediction (real-valued). Domains: 23 countries}
\end{center}
The task is to predict the real-valued asset wealth index of an area, given its satellite imagery.
Since the number of domains considered here is large (23 countries), instead of looping over all $S$ domains in each inner-loop, we sample $N<<S$ domains in each iteration and perform inner-loop updates using minibatches from these domains only to speed up computation.
For this dataset we choose $N=5$ by hyper-parameter search.

\textbf{Evalutaion:} \emph{Pearson Correlation (r).} Following the practice in \textsc{Wilds} benchmark, we compare the results by computing Pearson correlation (r) between the predicted and ground-truth asset index over 3 random seed runs.

\textbf{Results:}
We train the model using a ResNet-18 \citep{resnet} backbone.
See \Cref{tab:poverty_results}.

We see that Fish obtains the highest test performance, with the same validation performance as the best baseline. The performance is more stable between validation and test, and the standard deviation is smaller than for the baselines.
We also report the results of \gls{ERM} models trained in our environment as ``ERM (ours)'', which shows similar performance to the canonical results reported in the \textsc{Wilds} benchmark itself (``ERM'').

\subsection{\textsc{Camelyon17-Wilds}}
\label{sec:camelyon-wilds}
\begin{center}
    \emph{Task: Tumor detection (2 classes). Domains: 5 hospitals}
\end{center}
The \textsc{Camelyon17-Wilds} dataset contains 450,000 lymph-node scans from 5 hospitals.
Due to the size of the dataset, instead of training with Fish from scratch, we pre-train the model with ERM using the recommended hyper-parameters in \citet{wilds}, and fine-tune with Fish.
For this dataset, we find that Fish performs the best when starting from a pretrained model that has not yet converged, achieving much higher accuracy than the \gls{ERM} model.
we provide an ablation study on this in \Cref{sec:pretrained}.

\textbf{Evaluation:} \emph{Average accuracy.}
We evaluate the average accuracy of this binary classification task.
Following \citet{wilds}, we show the mean and standard deviation of results over 10 random seeds runs.
The number of random seeds required here is greater than other \textsc{Wilds} datasets due to the large variance observed in results.
Note that these random seeds are not only applied during the fine-tuning stage, but also to the pretrained models to ensure a fair comparison.

\textbf{Results:}
Following the practice in \textsc{Wilds}, we adopt DenseNet-121's \citep{dense} architecture for models trained on this dataset.
See results in \Cref{tab:camelyon_results}.

The results show that Fish significantly outperforms all baselines -- its test accuracy surpasses the best performing baseline by $6\%$.
Also note that for all other baselines, there is a large gap between validation and test accuracy ($11\% \~ 27\%$).
This is because \textsc{Wilds} chose the hospital that is the most difficult to generalize to as the test split to make the task more challenging.
Surprisingly, as we can observe in \Cref{tab:camelyon_results}, the discrepancy between test and validation accuracy of Fish is quite small ($3\%$).
The fact that it is able to achieve a similar level of accuracy on the worst-performing domain further demonstrates that Fish does not rely on domain-specific information, and instead makes predictions using the invariant features across domains.

\begin{minipage}{0.48\linewidth}
\vspace{-15pt}
\begin{table}[H]
\centering
\caption{Results on \textsc{FMoW-Wilds}.}
\scalebox{0.7}{
\begin{tabular}{lccccc}    \toprule
    \multirow{2}{*}{Method} & \multicolumn{2}{c}{Val. Accuracy ($\%$)} & & \multicolumn{2}{c}{Test Accuracy ($\%$)}  \\
    \cmidrule{2-3} \cmidrule{5-6}
     & Average & Worst & & Average & \textbf{Worst} \\
    \midrule
    {\textbf{Fish}} & 57.3 \std{0.01} & 49.5 \std{0.44} && 51.8 \std{0.12} & \textbf{34.3 \std{0.61}}\\
    {ERM} & 59.7 \std{0.14} & 48.2 \std{2.05} && {53.1 \std{0.25}} & 31.7 \std{1.01}\\
    {ERM} (ours) & 59.9 \std{0.22} & 47.1 \std{1.21} && {52.9 \std{0.18}} & 30.9 \std{1.53}\\
    {IRM} & 57.2 \std{0.01} & 47.4 \std{2.36} && 50.9 \std{0.32} & 31.0 \std{1.15}\\
    {Coral} & 56.7 \std{0.06} & 46.8 \std{1.18} && 50.5 \std{0.30} & 30.5 \std{0.70}\\
    \bottomrule
\end{tabular}\label{tab:fmow_results}}
\end{table}
\end{minipage}\hspace{6pt}\begin{minipage}{0.48\linewidth}
\vspace{-15pt}
\begin{table}[H]
\centering
\caption{\centering Results on \textsc{CivilComments-Wilds}.}\vspace{3pt}
\scalebox{0.7}{
\begin{tabular}{lccccc}
    \toprule
    \multirow{2}{*}{Method} & \multicolumn{2}{c}{Val. Accuracy ($\%$)} & & \multicolumn{2}{c}{Test Accuracy ($\%$)}  \\
    \cmidrule{2-3} \cmidrule{5-6}
     & Average & {Worst} & & Average & \textbf{Worst} \\
    \midrule
    \textbf{Fish} & 91.8 \std{0.2} & 75.3 \std{0.3} && 91.4 \std{0.3} & \textbf{74.2 \std{0.5}}\\
    Group DRO & 89.6 \std{0.3} & 68.7 \std{1.0} && 89.4 \std{0.3} & 70.4 \std{2.1}\\
    Reweighted & 89.1 \std{0.3} & 67.9 \std{1.2} && 88.9 \std{0.3} & 67.3 \std{0.1} \\
    \gls{ERM} & 92.3 \std{0.6} & 53.6 \std{0.7} && {92.2 \std{0.6}} & 58.0 \std{1.2}\\
    \gls{ERM} (ours) & 92.1 \std{0.5} & 54.1 \std{0.4} && {92.5 \std{0.3}} & 58.1 \std{1.7}\\
    \bottomrule
\end{tabular}\label{tab:civil_results}}
\end{table}
\end{minipage}

\subsection{\textsc{FMoW-Wilds}}
\label{sec:fmow-wilds}
\begin{center}
    \emph{Task: Infrastructure classification (62 classes). Domains: 80 (16 years x 5 regions)}
\end{center}
Similar to \textsc{Camelyon17-Wilds}, since the number of domains is large, we sample $N=5$ domains for each inner-loop.
To speed up computation, we also use a pretrained \gls{ERM} model and fine-tune with Fish; different from \Cref{sec:camelyon-wilds}, we find the best-performing models are acquired when using converged pretrained models (see details in \Cref{sec:pretrained}).

\textbf{Evaluation:}
\emph{Average $\&$ worst-region accuracies}.
Following \textsc{Wilds}, the average accuracy evaluates the model's ability to generalize over years, and the worst-region accuracy measures the model's performance across regions under a time shift.
We report results using 3 random seeds.

\textbf{Results:}
Following \citet{wilds}, we use a DenseNet-121 pretrained on ImageNet for this dataset.
Results in \Cref{tab:fmow_results} show that Fish has the highest worst-region accuracy on both test and validation sets.
It ranks second in terms of average accuracy, right after \gls{ERM}.
Again, Fish's performance is notably stable with the smallest standard deviation across all metrics compared to baselines.

\subsection{\textsc{CivilComments-Wilds}}
\label{sec:civilcomments-wilds}
\begin{center}
    \emph{Task: Toxicity detection in online comments (2 classes). Domains: 8 demographic identities.}
\end{center}
The \textsc{CivilComments-Wilds} contains 450,000 comments collected from online articles, each annotated for toxicity and the mentioning of demographic identities.
Again, we use ERM pre-trained model to speed up computation, and sample $N=5$ domains for each inner-loop.

\textbf{Evaluation:} \emph{Worst-group accuracy}.
To study the bias of annotating comments that mentions demographic groups as toxic, the \textsc{Wilds} benchmark proposes to evaluate the model's performance by doing the following:
1) Further separate each of the 8 demographic identities into 2 groups by toxicity -- for example, separate \emph{black} into \emph{black, toxic} and \emph{black, not toxic};
2) measure the accuracies of these $8\times2=16$ groups and use the lowest accuracy as the final evaluation of the model.
This metric is equivalent to computing the sensitivity and specificity of the classifier on each demographic identity, and reporting the worse of the two metrics over all domains.
Good performance on the group with the worst accuracy implies that the model does not tend to use demographic identity as an indicator of toxic comments.

Again, following \citet{wilds} we report results of 3 random seed runs.

\textbf{Results:}
We compare results to the baselines used in the \textsc{Wilds} benchmark over 3 random seed runs in \Cref{tab:civil_results}.
All models are trained using BERT \citep{bert}.

The results show that Fish outperforms the best baseline by $4\%$ and $7\%$ on the test and validation set's worst-group accuracy respectively, and is competitive in terms of average accuracy with \gls{ERM} (within standard deviation).
The strong performance of Fish on worst-group accuracy suggests that the model relies the least on demographic identity as an indicator of toxic comments compared to other baselines.
\gls{ERM}, on the other hand, has the highest average accuracy and the lowest worst-group accuracy.
This indicates that it achieves good average performance by leveraging the spurious correlation between toxic comments and the mention of certain demographic groups.

Note that different from all other datasets in \textsc{Wilds} that focus on \emph{pure domain generalization} (i.e, no overlap between domains in train and test splits), \textsc{CivilComments-Wilds} is a \emph{subpopulation shift} problem, where the domains in test are a subpopulation of the domains in train.
As a result, the baseline models used in \textsc{Wilds} for this dataset are different from the methods used in all other datasets, and are tailored to avoiding systematic failure on data from minority subpopulations.
Fish works well in this setting too without any changes or special sampling strategies (such as $*$ and $+$ in \Cref{tab:civil_results}).
This further demonstrates the good performance of our algorithm on different domain generalization scenarios.

\subsection{\textsc{iWildCam-Wilds}}
\label{sec:iwildcam-wilds}

\begin{center}
    \emph{Task: Animal species (186 classes). Domains: 324 camera locations.}
\end{center}
The dataset consists of over 200,000 photos of animal in the wild, using stationary cameras across 324 locations. 
Classifying animal species from these heat or motion-activated photos is especially challenging: methods can easily rely on the background information of photos from the same camera setup. 
Fish models are pretrained with ERM till convergence, and for each inner loop we sample from $N=10$ domains.

\textbf{Evaluation:} \emph{Macro F1 score}. Across the 186 class labels, we report average accuracy and both weighted and macro F1 scores (F1-w and F1-m, respectively, in \Cref{tab:iwc_results}).
We run $3$ random seeds for each model.

\textbf{Results:} All models reported in \Cref{tab:camelyon_results} are trained using a ResNet-50. We find Fish to outperform baselines on both test accuracy and weighted F1, with a $1\%$ improvement on both metrics over the best performing model (\gls{ERM}). However, this comes at the cost of lower macro F1 score, where Fish performs $1\%$ worse than \gls{ERM} models that we trained and $3\%$ than the \gls{ERM} reported in \textsc{Wilds}. This suggests that Fish is less good at classifying rarer species, however the overall accuracy on the test dataset is improved.

Although Fish did not outperform the ERM baseline on the primary evaluation metric proposed in \citet{wilds},  we found the improvement of Fish in both accuracy and weighted F1 to be robust across a range of hyperparameters.
See more details on this in \Cref{sec:hyper_app}.

\subsection{\textsc{Amazon-Wilds}}
\label{sec:amazon-wilds}
\begin{center}
    \emph{Task: Sentiment analysis (5 classes). Domains: 7,676 Amazon reviewers.}
\end{center}
The dataset contains 1.4 million customer reviews on Amazon from 7,676 customers, and the task is to predict the score (1-5 stars) given the review.
Similarly, we pretrained the model with ERM till convergence, and due to the large number of domains ($S=5008$ in train) we sample $N=5$ reviewers for each inner loop.

\textbf{Evaluation:} \emph{10th percentile accuracy}. Reporting the accuracy of the 10th percentile reviewer helps us assess whether the model performance is consistent across different reviewers.
The results in \Cref{tab:amazon_results} are reported over 3 random seeds.

\textbf{Results:} The model is trained using \textsc{BERT} \citep{bert} backbone. While Fish has lower average accuracy compared to ERM, its 10th percentile accuracy matches that of \gls{ERM}, outperforming all other baselines.

\vspace{-20pt}
\begin{center}
\begin{table}[H]
\centering
\caption{Results on \textsc{iWildCam-Wilds}.}
\scalebox{0.7}{
\begin{tabular}{lcccccccc}
    \toprule
    \multirow{2}{*}{Method} & \multicolumn{3}{c}{Validation} & &  \multicolumn{3}{c}{Test} \\
    \cmidrule{2-4} \cmidrule{6-8}
     & Acc. (\%) & Weighted F1 & Macro F1 & & Acc. (\%) & Weighted F1 & \textbf{Macro F1}  \\
    \midrule
    Fish & 58.0 \std{0.2} & 56.5 \std{0.3} & 25.8 \std{0.5} & & {63.2 \std{0.7}} & {64.0 \std{0.5}} & 24.2 \std{0.9} \\
    Coral & - & - & - & & 62.5 \std{1.7} & - & 26.3 \std{1.4}\\
    \textbf{ERM} & - & -  & - & & 62.9 \std{0.5} & - & \textbf{{27.8 \std{1.3}}}\\
    ERM (ours) & 55.8 \std{0.2} & 54.6 \std{0.5} & 27.7 \std{1} & & 63.0 \std{0.6} & 62.4 \std{0.2} & 25.1 \std{0.2} \\
    \bottomrule
\end{tabular}\label{tab:iwc_results}}
\end{table}
\begin{table}[H]
\centering
\caption{Results on \textsc{Amazon-Wilds}.}
\scalebox{0.7}{
\begin{tabular}{lccccc}    \toprule
    \multirow{2}{*}{Method} & \multicolumn{2}{c}{Val. Accuracy ($\%$)} & & \multicolumn{2}{c}{Test Accuracy ($\%$)}  \\
    \cmidrule{2-3} \cmidrule{5-6}
     & Average & {10-th per.} & & Average & \textbf{10-th per.} \\
    \midrule
    {\textbf{Fish}} & 73.8 \std{0.1} & 57.3 \std{0.0} && 72.9 \std{0.2} & \textbf{56.0 \std{0.0}}\\
    {\textbf{ERM}} & 74.3 \std{0.0} &  57.3 \std{0.0} && 73.5 \std{0.1} & \textbf{56.0 \std{0.0}} \\
    {\textbf{ERM} (ours)} & 74.0 \std{0.0} & 57.3 \std{0.0} && 73.3 \std{0.1} & \textbf{56.0 \std{0.0}}\\
    {IRM} & 73.6 \std{0.2} & 56.4 \std{0.8} && 73.3 \std{0.2} & 55.1 \std{0.8}\\
    {Reweighted} & 69.6 \std{0.0} &  53.3 \std{0.0} &&  69.2 \std{0.1} & 52.4 \std{0.8}\\
    \bottomrule
\end{tabular}\label{tab:amazon_results}}
\end{table}
\end{center}

\section{Hyperparameters}\label{sec:hyper_app}
In \Cref{tab:erm_hyper} we list the hyperparameters we used to train ERM.
The same hyperparameters were used for producing ERM baseline results and as pretrained models for Fish.
In \texttt{val. metric} we report the metric on validation set that is used for model selection, and in \texttt{cut-off} we specify when to stop training when using ERM to generate pretrained models.
\begin{table*}[ht!]
\centering
\caption{Hyperparameters for ERM. We follow the hyperparameters used in \textsc{Wilds} benchmark. Note that we did not use a pretrained model for  \textsc{Poverty}, therefore its cut-off condition is not reported.}\label{tab:erm_hyper}
\scalebox{0.8}{
\begin{tabular}{lccccccc}
    \toprule
    Dataset & Model & Learning rate & Batch size & Weight decay & Optimizer & Val. metric & Cut-off \\
    \midrule
    \textsc{Camelyon17} & Densenet-121 & 1e-3 & 32 & 0 & SGD & acc. avg. & iter 500\\
    \textsc{CivilComments} & BERT & 1e-5 & 16 & 0.01 & Adam & acc. wg. & Best val. metric\\
    \textsc{FMoW} & Densenet-121 & 1e-4 & 64 & 0 & Adam & acc. avg. & Best val. metric\\
    \textsc{iWildCam} & Resnet-50 & 1e-4 & 16 & 0 & Adam & F1-macro (all) & Best val. metric\\
    \textsc{Poverty} & Resnet-18 & 1e-3 & 64 & 0 & Adam & Pearson (r) & -\\
    \textsc{Amazon} & BERT & 2e-6 & 8 & 0.01 & Adam & 10th percentile acc. & -\\
    \bottomrule
\end{tabular}}
\end{table*}

In \Cref{tab:fish_hyper} we list out the hyperparameters we used to train Fish.
Note that we train Fish using the same model, batch size, val metric and optimizer as ERM -- these are not listed in \Cref{tab:fish_hyper} to avoid repetitions.
Weight decay is always set as 0.

\begin{table*}[ht!]
\centering
\caption{Hyperparameters for Fish. }\label{tab:fish_hyper}
\scalebox{0.8}{
\begin{tabular}{llcccc}
    \toprule
    Dataset  & Group by &  $\alpha$ &  $\epsilon$ & $\#$ domains & Meta steps \\
    \midrule
    \textsc{Camelyon17} & Hospitals &  1e-3  & 0.01 & 3 & 3 \\
    \textsc{CivilComments} & Demographics $\times$ toxicity & 1e-5 & 0.05 & 16 & 5 \\
    \textsc{FMoW} & time $\times$ regions &  1e-4 & 0.01 & 80 & 5\\
    \textsc{iWildCam} & Trap locations & 1e-4 & 0.01 & 324 & 10\\
    \textsc{Poverty} & Countries & 1e-3 & 0.1 & 23 & 5\\
    \textsc{Amazon} & Reviewers & 2e-6 & 0.01 & 7,676 & 5\\
    \bottomrule
\end{tabular}}
\end{table*}

\section{Ablation Studies on Pre-trained Models}\label{sec:pretrained}
In this section we perform ablation study on the convergence of pretrained ERM models.
We study the performance of Fish with the following three configurations of pretrained ERM models:
\begin{enumerate}
    \item Model is trained on $10\%$ of the data (epoch 1);
    \item Model is trained on $50\%$ of the data (epoch 1);
    \item Model at convergence.
\end{enumerate}
By comparing the results between these three settings, we demonstrate how the level of convergence affects the Fish's training performance.
See results in \Cref{tab:pretrained_results}.
Note that \textsc{Poverty} is excluded here because the dataset is small enough that we are able to train Fish from scratch.

\begin{table}[H]
\centering
\caption{Ablation study on pretrained ERM models. }\label{tab:pretrained_results}
\scalebox{0.7}{
\begin{tabular}{lcccc}
    \toprule
    \multirow{2}{*}{Model}  & \textsc{FMoW} & \textsc{Camelyon17} & \textsc{iWildCam} & \textsc{CivilComments} \\
    \cmidrule{2-5}
    & Test Avg Acc & Test Avg Acc & Test Macro F1 & Test Worst Acc\\ 
    \midrule
    $10\%$ data & 21.7 \std{2.5} & 79.1 \std{12.3} & 13.7 \std{0.5} & 71.8 \std{1.3}\\
    $50\%$ data & 31.0 \std{0.8} & 64.6 \std{12.3} & 19.0 \std{0.06} & 74.2 \std{0.5}\\
    Converged & 32.7 \std{1.2} & 63.5 \std{8.2} & 23.7 \std{0.9} & 73.8 \std{1.8}\\
    \bottomrule
\end{tabular}}
\end{table}

We see that \textsc{CivilComments} sustain good performance using pretrained models at different convergence levels.
\textsc{FMoW} and \textsc{iWildCam} on the other hand seems to have strong preference towards converged model, and the results worsen as the amount of data seen during training goes down.
\textsc{Camelyon17} achieves the best performance when only $10\%$ of data is seen, and the test accuracy decreases while training with models with higher level of convergence.

\section{Ablation Studies on hyperparameters}\label{sec:abla_app}
\begin{figure*}
    \centering
    \begin{subfigure}{0.3\linewidth}
        \centering
        \includegraphics[width=\linewidth, trim={0.4cm 0.2cm 1cm 1cm}, clip]{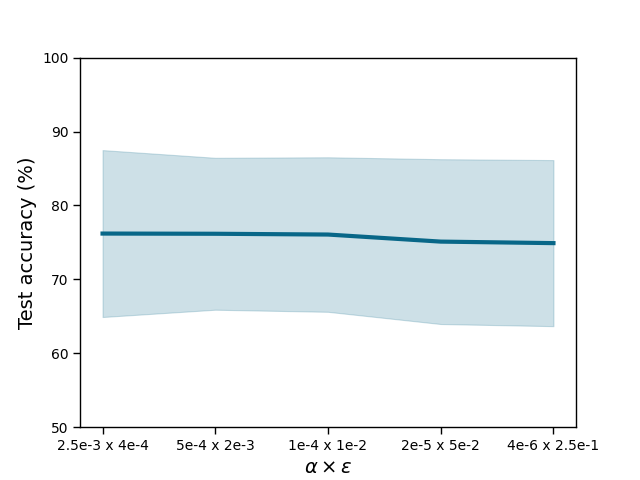}
        \caption{\textsc{Camelyon17}}
    \end{subfigure}
    \begin{subfigure}{0.3\linewidth}
        \centering
        \includegraphics[width=\linewidth, trim={0.6cm 0.2cm 1cm 1cm}, clip]{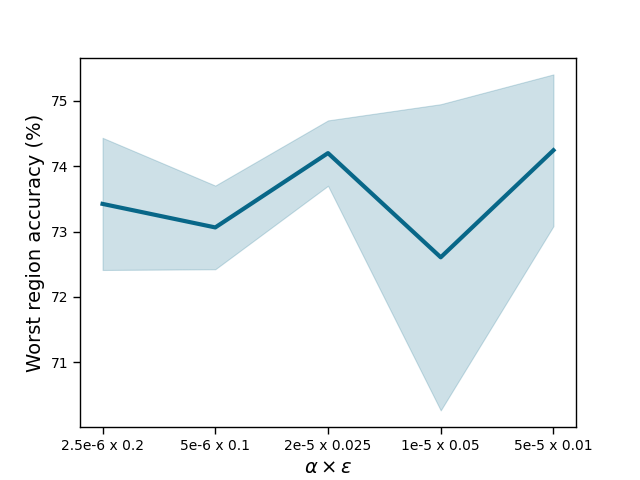}
        \caption{\textsc{CivilComments}}
    \end{subfigure}
    \begin{subfigure}{0.3\linewidth}
        \centering
        \includegraphics[width=\linewidth, trim={0.6cm 0.2cm 1cm 1cm}, clip]{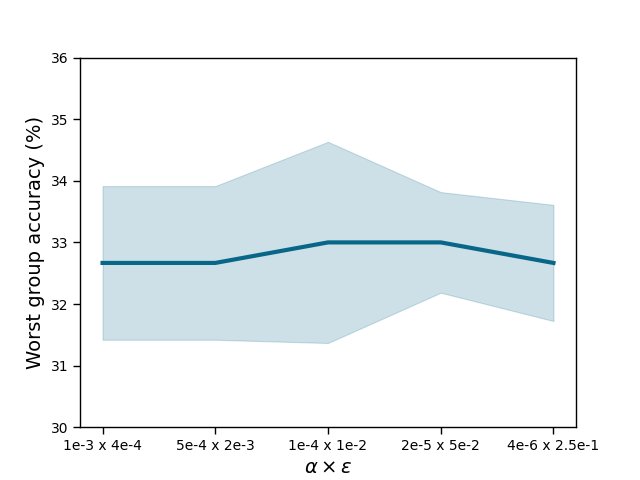}
        \caption{\textsc{FMoW}}
    \end{subfigure}\\
    \begin{subfigure}{0.3\linewidth}
        \centering
        \includegraphics[width=\linewidth, trim={0.2cm 0.2cm 1cm 1cm}, clip]{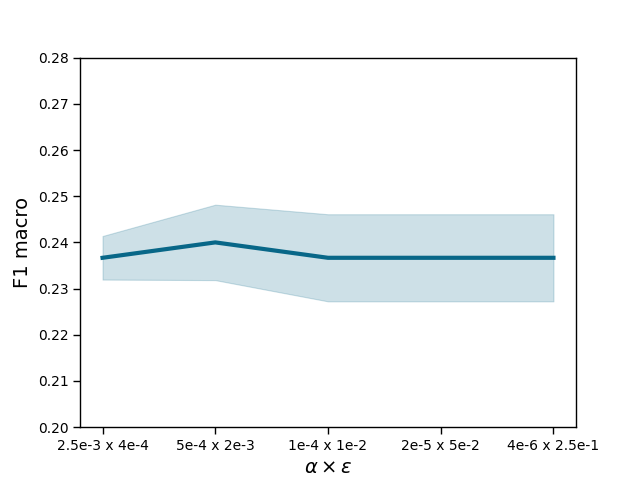}
        \caption{\textsc{iWildCam}}
    \end{subfigure}
    \begin{subfigure}{0.3\linewidth}
        \centering
        \includegraphics[width=\linewidth, trim={0.6cm 0.2cm 1cm 1cm}, clip]{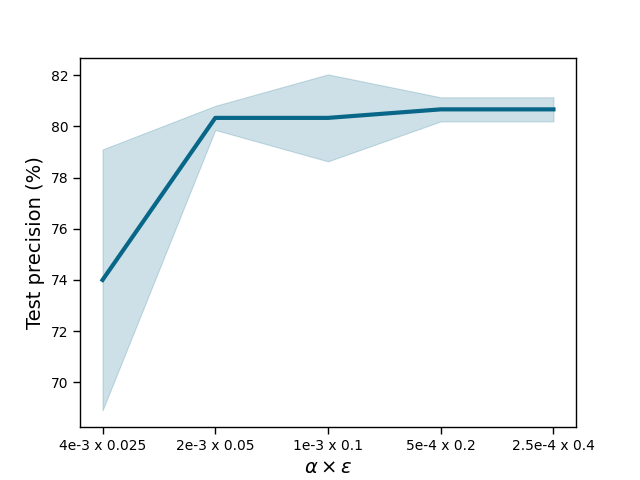}
        \caption{\textsc{Poverty}}
    \end{subfigure}
    \caption{Ablation studies on $\alpha$ and $\epsilon$. Note that $\alpha \times \epsilon$ remains constant in all experiments, and the midpoint of each plot is the hyperparameter we chose to use to report our experiment results.}\label{fig:alpha_epsilon}
\end{figure*}

\paragraph{$\alpha$ and $\epsilon$}
We study the effect of changing Fish's inner loop learning rate $\alpha$ and outer loop learning rate $\epsilon$.
To make the comparisons more meaningful, we keep $\alpha \cdot \epsilon$ constant while changing their respective values.
See results in \Cref{fig:alpha_epsilon}.

\paragraph{Meta steps $N$}
For most of the datasets we studied (all apart from \textsc{Camelyon17} where $T=3$) we sample a $N$-sized subset of all $T$ domains available for training (see \Cref{tab:fish_hyper} for $T$ of each dataset).
Here we study when $N=5, 10, 20$.

\begin{table}[H]
\centering
\caption{Ablation study on meta steps $N$. }\label{tab:s_econstant}
\scalebox{0.8}{
\begin{tabular}{lcccc}
    \toprule
    \multirow{2}{*}{$N$}  & \textsc{FMoW} & \textsc{Poverty} & \textsc{iWildCam} & \textsc{CivilComments} \\
    \cmidrule{2-5}
    & Test Avg Acc & Test Pearson r & Test Macro F1 & Test Worst Acc\\ 
    \midrule
    $5$ & 33.0 \std{1.6} & 80.3 \std{1.7} & 23.7 \std{0.9} & 74.3 \std{1.5} \\
    $10$ & 32.7 \std{1.2} & 80.0 \std{0.8} & 23.7 \std{0.5} & 73.4 \std{1.0} \\
    $20$ & 33.3 \std{2.1} & 77.7 \std{2.1} & 23.7 \std{0.9} & 72.6 \std{2.3} \\
    \bottomrule
\end{tabular}}
\end{table}

In general altering these hyperparameters don't have a huge impact on the model's performance, however it does seem thet when $N=20$ the performance on some datasets (\textsc{Poverty} and \textsc{CivilComments}) degrade slightly.

\section{Tracking gradient inner product}\label{sec:track_grad_app}

\begin{figure*}[ht!]
\centering
\begin{subfigure}{0.32\linewidth}
    \centering
    \includegraphics[width=\linewidth]{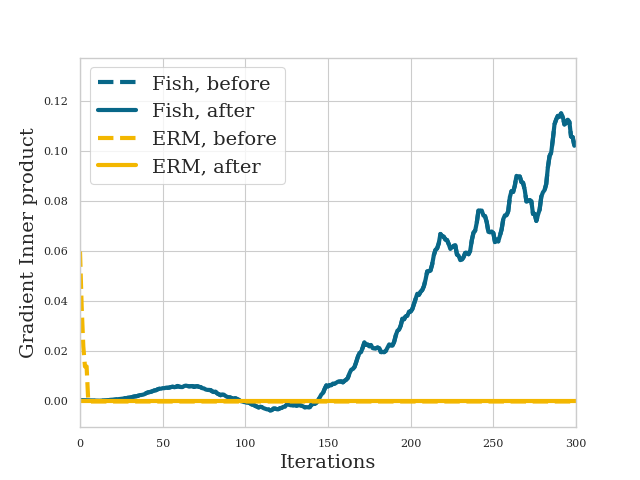}
    \caption{\textsc{CdSprites-N}} \label{fig:gcdsprites}
\end{subfigure}
\begin{subfigure}{0.32\linewidth}
    \centering
    \includegraphics[width=\linewidth]{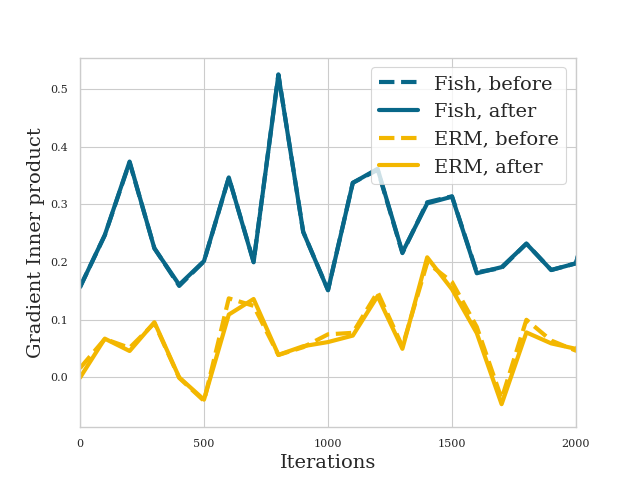}
    \caption{\textsc{Camelyon17}} \label{fig:gcam}
\end{subfigure}
\begin{subfigure}{0.32\linewidth}
    \centering
    \includegraphics[width=\linewidth]{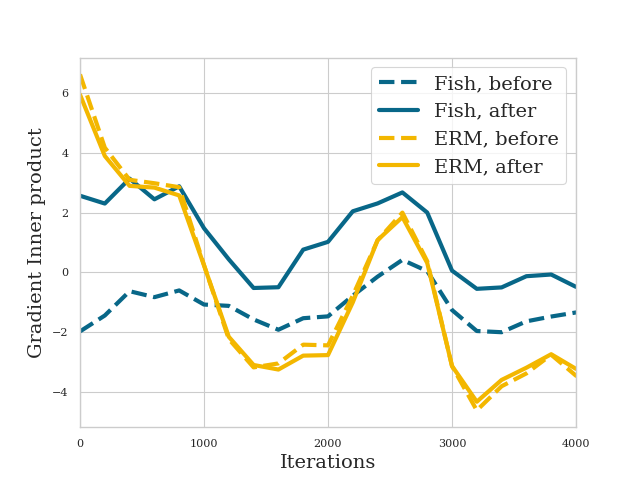}
    \caption{\textsc{CivilComments}} \label{fig:gcivil}
\end{subfigure}
\begin{subfigure}{0.32\linewidth}
    \centering
    \includegraphics[width=\linewidth]{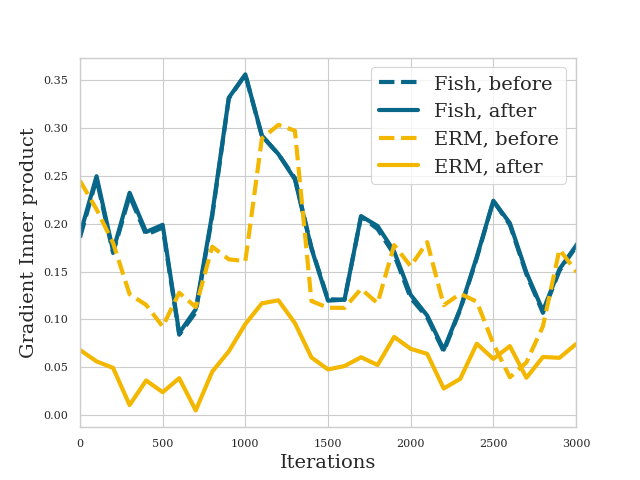}
    \caption{\textsc{FMoW}} \label{fig:gfmow}
\end{subfigure}
\begin{subfigure}{0.32\linewidth}
    \centering
    \includegraphics[width=\linewidth]{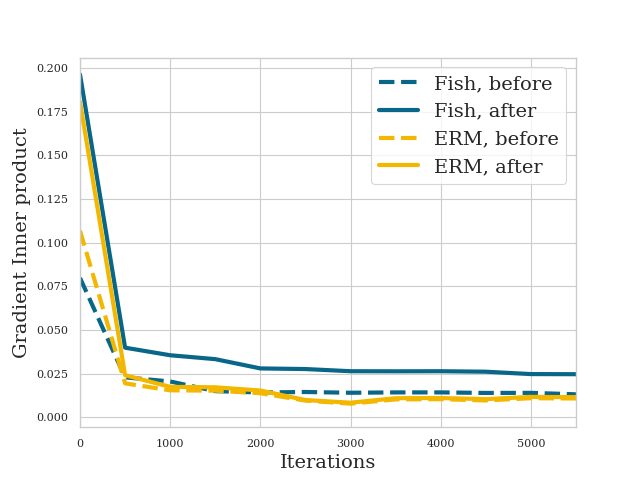}
    \caption{\textsc{Poverty}} \label{fig:gpoverty}
\end{subfigure}
\begin{subfigure}{0.32\linewidth}
    \centering
    \includegraphics[width=\linewidth]{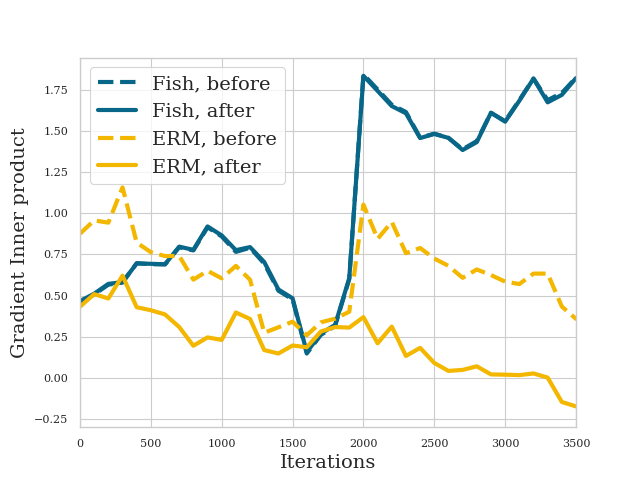}
    \caption{\textsc{iWildCam}} \label{fig:giwc}
\end{subfigure}
\caption{Gradient inner product values during the training for \textsc{CdSprites-N} (N=15) and 5 different \textsc{Wilds} datasets.}\label{fig:gradprog_wilds}
\end{figure*}
In \Cref{fig:gradprog_wilds}, we demonstrate the progression of inter-domain gradient inner products during training using different objectives.
We train both {\color{c1}\textbf{Fish}} (blue) and {\color{c3}\textbf{ERM}} (yellow) untill convergence while recording the normalized gradient inner products (i.e. cosine similarity) between minibatches from different domains used in each inner-loop.
The gradient inner products are computed both before (dotted) and after (solid) the model's update. 
To ensure a fair comparison, we use the exact same sequence of data for Fish and ERM (see \Cref{sec:trackinggradipm} for more details).

Inevitably, the gradient inner product trends differently for each dataset since the data, types of domain splits and the choice of architecture are very different.
In fact, the plot for \textsc{CdSprites-N} and \textsc{Poverty} are significantly different from others, with a dip in gradient inner product at the beginning of training -- this is because these are the two datasets that we train from scratch.
On all other datasets, the gradient inner products are recorded when fine-tuning with Fish.

Despite their differences, there are some important commonalities between these plots: if we compare the pre-update (dotted) to post-update (solid) curves, we can see that ERM updates often result in the decrease of gradient inner product, while for Fish it can either increase significantly (\Cref{fig:gcivil} and \Cref{fig:gpoverty}) or at least stay at the same level (\Cref{fig:gcdsprites}, \Cref{fig:gcam}, \Cref{fig:gfmow} and \Cref{fig:giwc}).
As a result of this, we can see that the post-update gradient inner product of Fish is always greater than ERM at convergence.

The observations here shows that Fish is effective in increasing/maintaining the level of inter-domain gradient inner product and sheds some lights on its efficiency on the datasets we studied.

\section{Algorithm for tracking gradient inner product}\label{sec:trackinggradipm}
To make sure that the gradients we record for ERM and Fish are comparable, we use the same sequence of $S$-minibatches to train both algorithms.
See \Cref{alg:tracking} for details.
\setlength{\textfloatsep}{1.5ex}
\begin{algorithm}[H]
    \begin{algorithmic}[1]
    \makeatletter
    \newcommand{\ourlinenumber}[1]{
    \let\old@ALC@lno=\ALC@lno
    \renewcommand{\ALC@lno}{\raggedright\begin{small}#1\end{small}
    \let\ALC@lno=\old@ALC@lno}
    }
    \makeatother
    \STATE \textbf{function} $\texttt{\bfseries{GIP}}(\{\d_1, \d_2, \cdots, \d_N\}, \theta)$:
    \FOR{$\d_n \in \{\d_1, \d_2, \cdots, \d_N\}$}
        \STATE $\g_n = \partial l(\bm{d}_n;\theta)/\partial\theta$
    \ENDFOR
    \STATE $\bar{\bar{\bm{g}}}=\frac{1}{S(S-1)} \sum_{i,j\in S}^{i\neq j} \g_i\cdot \g_j$
    \STATE \textbf{return} $\bar{\bar{\bm{g}}}$
    \end{algorithmic}
\caption{Function \texttt{\bfseries{GIP}}}\label{alg:gip}
\end{algorithm}

\setlength{\textfloatsep}{1.5ex}
\begin{algorithm}[t]
    \begin{algorithmic}[1]
    \makeatletter
    \newcommand{\ourlinenumber}[1]{
    \let\old@ALC@lno=\ALC@lno
    \renewcommand{\ALC@lno}{\raggedright\begin{small}#1\end{small}
    \let\ALC@lno=\old@ALC@lno}
    }
    \makeatother
    \STATE Initialize Fish $\theta_f\gets \theta$, ERM $\theta_{e}\gets \theta$
    \FOR{i = $1, 2, \cdots$}
        \STATE \COMMENT{Get all minibatches}
        \FOR{$\D_n \in \{\D_1, \D_2, \cdots, \D_N\}$}
            \STATE Sample batch  $\bm{d}_n \sim \D_n$
        \ENDFOR
        \STATE \COMMENT{GradInnerProd before update}
        \STATE $\bar{\bar{\bm{g}}}_{Fb}=\texttt{\bfseries GIP}(\{\d_1, \d_2, \cdots, \d_N\}$, $\theta_f$)
        \STATE $\bar{\bar{\bm{g}}}_{Eb}=\texttt{\bfseries GIP}(\{\d_1, \d_2, \cdots, \d_N\}$, $\theta_e$)
        \STATE \COMMENT{Fish training}
        \STATE $\tilde{\theta} \gets \theta_f$
        \FOR{$\d_n \in \{\d_1, \d_2, \cdots, \d_N\}$}
            \STATE $\g_n = \partial l(\bm{d}_n;\tilde{\theta})/\partial\tilde{\theta}$
            \STATE Update $\tilde{\theta} \gets \tilde{\theta} - \alpha\g_n$
        \ENDFOR
        \STATE $\theta_f \gets \theta_f + \displaystyle \epsilon(\tilde{\theta}-\theta_f)$
        \STATE \COMMENT{Rearrange minibatches}
        \STATE $\d=\texttt{shuffle}(\texttt{concat}(\d_1, \d_2, \cdots, \d_N))$
        \STATE $\{\tilde{\d}_1, \tilde{\d}_2, \cdots, \tilde{\d}_N\}=\texttt{split}(\d)$
        \STATE \COMMENT{ERM training}
        \FOR{$\tilde{\d}_n \in \{\tilde{\d}_1, \tilde{\d}_2, \cdots, \tilde{\d}_N\}$}
            \STATE $\g_n = \partial l(\tilde{\d}_n;\theta_e)/\partial\theta_e$
            \STATE Update $\theta_e \gets \theta_e - \alpha\g_n$
        \ENDFOR
        \STATE \COMMENT{GradInnerProd after update}
        \STATE $\bar{\bar{\bm{g}}}_{Fa}=\texttt{\bfseries GIP}(\{\d_1, \d_2, \cdots, \d_N\}$, $\theta_f$)
        \STATE $\bar{\bar{\bm{g}}}_{Ea}=\texttt{\bfseries GIP}(\{\d_1, \d_2, \cdots, \d_N\}$, $\theta_e$)
    \ENDFOR
    \STATE \textbf{Return} $\bar{\bar{\bm{g}}}_{Fb}$, $\bar{\bar{\bm{g}}}_{Fa}$, $\bar{\bar{\bm{g}}}_{Eb}$, $\bar{\bar{\bm{g}}}_{Ea}$
    \end{algorithmic}
\caption{Algorithm of collecting gradient inner product $\bar{\bar{\bm{g}}}$ for \underline{\textbf{F}}ish and \underline{\textbf{E}}RM both \underline{\textbf{b}}efore and \underline{\textbf{a}}fter updates. See \texttt{\bfseries{GIP}} in \Cref{alg:gip}.}\label{alg:tracking}
\end{algorithm}


\end{document}